\documentclass[11pt, letterpaper]{article}
\usepackage[margin=1in]{geometry}

\usepackage{amsfonts, amsmath, amssymb, amsthm}
\usepackage{bbm}
\usepackage{color}
\usepackage{euscript}
\usepackage{graphicx}
\usepackage{hyperref}
\usepackage{mathrsfs} 
\usepackage{microtype}
\usepackage{tcolorbox}
\usepackage[normalem]{ulem}
\usepackage{wrapfig}

\def\ttt{\texttt}

\def\extraspacing{\vspace{2mm} \noindent}
\def\figcapup{\vspace{-1mm}}
\def\figcapdown{\vspace{-0mm}}

\def\vgap{\vspace{1mm}}

\def\tabpos{\hspace{4mm} \= \hspace{4mm} \= \hspace{4mm} \= \hspace{4mm} \=
\hspace{4mm} \= \hspace{4mm} \= \hspace{4mm} \= \hspace{4mm} \= \hspace{4mm}
\kill}
\newcommand{\mytab}[1]{\begin{tabbing}\tabpos #1\end{tabbing}}

\newtheorem{example}{Example}[section]

\newtheorem{theorem}{Theorem}

\newtheorem{lemma}[theorem]{Lemma}
\newtheorem{corollary}[theorem]{Corollary}
\newtheorem{proposition}{Proposition}

\newtheorem{problem}{Problem}

\newcommand{\boxminipg}[2]{\vspace{2mm} \begin{center}\fbox{\begin{minipage}{#1}#2\end{minipage}}\end{center} \vspace{2mm}}
\newcommand{\minipg}[2]{\vspace{2mm}  \begin{center}\begin{minipage}{#1}#2\end{minipage}\end{center} \vspace{2mm} }
\newcommand{\myitems}[1]{\begin{itemize}\setlength #1 \end{itemize}}

\newcommand{\myfigg}[2]{\begin{figure}\centering #1 \figcapup \caption{#2} \figcapdown \end{figure}}

\newcommand{\bm}[1]{\textrm{\boldmath${#1}$}}

\newcommand{\myeqn}[1]{\begin{eqnarray}#1\end{eqnarray}}

\newcommand{\set}[1]{\{#1\}}

\newcommand{\explain}[1]{(\textrm{#1})}

\def\mit{\mathit}
\def\eps{\epsilon}
\def\fr{\frac}
\def\-{\mbox{-}}

\def\real{\mathbb{R}}

\def\tO{\tilde{O}}

\def\lc{\lceil}

\def\rc{\rceil}

\def\nn{\nonumber}

\def\Pr{\mathbf{Pr}}
\DeclareMathOperator*\expt{\mathbf{E}}

\def\bigmid{\textrm{ $\Big|$ }}

\def\setm{\setminus}

\def\*{\star}

\DeclareMathOperator*{\polylog}{polylog}

\DeclareMathOperator*{\Log}{Log}

\usepackage{bbm}

\allowdisplaybreaks

\def\bad{\mit{bad}}
\def\bigmid {\, \Big\lvert \,}

\def\cost{\mit{cost}}

\def\det{\mit{det}}
\def\dis{\textrm{DIS}}
\def\dom{\succ}
\def\dombyeq{\preceq}
\def\domeq{\succeq}
\def\err{\mit{err}}
\def\errate{\mit{err}\textrm{-}\mit{rate}}
\def\good{\mit{good}}
\def\lab{\mit{label}}

\def\middle{\mit{mid}}
\def\mono{\mathbb{H}_\mit{mon}}
\def\neg{\mit{neg}}
\def\newstar{\circledast}
\def\pos{\mit{pos}}
\def\ran{\mit{ran}}
\def\rest{\mit{rest}}
\def\rpe{\mathrm{RPE}}
\def\siml{\mathit{sim}}

\def\starnum{\mathfrak{s}}
\def\totalcost{\mit{family\textrm{-}cost}}
\def\totalerr{\mit{family\textrm{-}err}}
\def\weight{\mit{weight}}
\def\werr{\mit{w}\textrm{-}\mit{err}}

\def\A{\mathcal{A}}

\def\D{\mathcal{D}}

\def\H{\mathcal{H}}
\def\T{\mathcal{T}}
\def\P{\mathcal{P}}
\def\Q{\mathcal{Q}}

\def\bbA{\mathbb{A}}
\def\bbH{\mathbb{H}}

\def\bbP{\mathbb{P}}

\def\LL{\EuScript{L}}

\def\TT{\EuScript{T}}

\def\EEE{\mathscr{E}}

\def\TTT{\mathscr{T}}

\def\rev{}
\def\revv{}
\def\revvv{}

\title{Monotone Classification with Relative Approximations}

\author{Yufei Tao \\[2mm]
  CUHK \\
  taoyf@cse.cuhk.edu.hk
}

\begin{document}

\maketitle

\begin{abstract}
    In monotone classification, the input is a multiset $P$ of points in $\real^d$, each associated with a hidden label from $\{-1, 1\}$. The goal is to identify a monotone classifier $h: \real^d \rightarrow \set{-1, 1}$ with a small {\em error}, measured as the number of points $p \in P$ whose labels differ from $h(p)$. The {\em cost} of an algorithm is the number of point labels it chooses to reveal. This article \rev{studies} the minimum cost required to find a monotone classifier whose error is at most $(1 + \eps) \cdot k^*$ where $\eps \ge 0$, and $k^*$ is the error of an optimal monotone classifier. In other words, the algorithm aims for a \rev{{\em relative} approximation}, allowing the error to exceed the optimum by at most a factor of $1+\eps$. We present nearly matching upper and lower bounds for the entire range of $\eps$. Previous work could guarantee only an \rev{additive approximation} to the optimal error.
\end{abstract}

\vspace{10mm}

Preliminary versions of this article appeared in PODS'18 and PODS'21. 


\thispagestyle{empty}

\newpage

\setcounter{page}{1}

\section{Introduction} \label{sec:intro}

This article presents a systematic study of monotone classification, aiming to determine the minimum label-discovery overhead to guarantee a classification error that is higher than the optimum by at most a relative factor. We begin by defining the problem and explaining its practical motivations. We then present our findings and discuss their significance in relation to previous results. Finally, we provide an overview of our algorithmic techniques.

\subsection{Problem Definitions} \label{sec:intro:prob}


\noindent {\bf Math Conventions.} For an integer $x \ge 1$, the notation $[x]$ represents the set $\set{1, 2, ..., x}$. Given two non-negative real values $x$ and $y$, we use the notation $\rev{x \le_{1+\eps} y}$ to represent the condition $x \le (1+\eps) y$ where  $\eps \ge 0$. Given a real value $x \ge 1$, we use $\Log x$ as a short form for $\log_2 (1 + x)$. For any real value $x$, the expression $\exp(x)$ denotes $e^x$. Given a predicate $Q$, the notation $\mathbbm{1}_Q$ equals 1 if $Q$ holds or 0 otherwise.

\vgap

Given a point $p \in \real^d$ for some dimensionality $d \ge 1$, the notation $p[i]$ represents the coordinate of $p$ on dimension $i \in [d]$. A point $p \in \real^d$ is said to {\em dominate} a point $q \in \real^d$ if $p \ne q$ and the condition $p[i] \ge q[i]$ holds for all $i \in [d]$. We use $p \dom q$ to indicate ``$p$ dominating $q$'' and $p \domeq q$ to indicate ``$p = q$ or $p \dom q$''. If $p \domeq q$, we may also write $q \dombyeq p$.

\vgap

\extraspacing {\bf Monotone Classification.} The input is a multiset $P$ of $n$ points in $\real^d$ for some integer $d \ge 1$. The multiset may contain repeated points, that is, distinct elements with identical coordinates. Each element $p \in P$ carries a label from $\set{-1, 1}$, which is represented as $\lab(p)$.

\vgap

A {\em classifier} is a function $h: \real^d \rightarrow \set{-1, 1}$. We say that $h$ {{\em correctly classifies}} an element $p \in P$ if $h(p) = \lab(p)$, or {\em misclassifies} $p$, otherwise. The {\em error} of $h$ on $P$ is defined as
\myeqn{
    \err_P(h) &=& \sum_{p \in P} \mathbbm{1}_{h(p) \neq \lab(p)} \label{eqn:err-classifier}
}
that is, the number of elements in $P$ misclassified by $h$. A classifier $h$ is {\em monotone} if $h(p) \ge h(q)$ holds for any two points $p, q \in \real^d$ satisfying $p \dom q$.

\vgap

Define
\myeqn{
    \mono &=& \text{the set of all monotone classifiers} \label{eqn:monoclass} \\
    k^* &=&
    \min_{h \in \mono} \err_P(h). \label{eqn:k-star}
}
We call $k^*$ the \rev{\em optimal monotone error} of $P$. A classifier $h \in \mono$ is
\myitems{
    \item {{\em optimal}} on $P$ if $\err_P(h) = k^*$;

    \item {\em $c$-approximate} on $P$ if $\err_P(h) \le c \cdot k^*$ where $c \ge 1$ is the {{\em approximation ratio}} of $h$.
}

\myfigg{
    \hspace{20mm}
    \includegraphics[height=55mm]{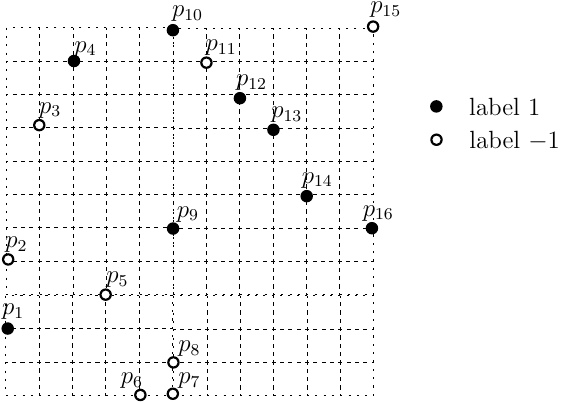}
}{An input set $P$ for Problem 1 \label{fig:intro-ds}}

\begin{example}
    Figure~\ref{fig:intro-ds} shows a 2D input $P$ where a black (resp., white) point has label 1 (resp., $-1$). Consider the monotone classifier $h$ that maps (i) all the black points to 1 except $p_1$, and (ii) all the white points to $-1$ except $p_{11}$ and $p_{15}$. Thus, $\err_P(h) = 3$. No other monotone classifier has a smaller error on $P$, and hence $k^* = 3$. Consider the classifier $h^\pos$ that maps the entire $P$ to $1$. \rev{It can be verified} that $\err_P(h^\pos) = \rev{8}$; \rev{hence,} $h^\pos$ is $\rev{(8/3)}$-approximate on $P$. 
\end{example}

The goal of an algorithm $\A$ is to find a classifier from $\mono$ whose error on $P$ is small. In the beginning, the labels of the elements in $P$ are hidden. There exists an {\em oracle} that $\A$ can query for labels. Specifically, in each {\em probe}, the algorithm selects an element $p \in P$ and receives $\lab(p)$ from the oracle. Its {\em cost} is the total number of elements probed.

\vgap

In one extreme, by probing the whole $P$, \rev{$\A$ pays a cost of $n$}, after which it can find an optimal monotone classifier on $P$ with \rev{polynomial-time} computation\footnote{Once all the point labels are available, an optimal monotone classifier can be found in time polynomial in $n$ and $d$; see \cite{ahkw06,s13b}.}. In the other extreme, by probing nothing, \rev{$\A$ pays a cost of 0} but has to return a classifier based purely on the point coordinates; such a classifier could be very erroneous on $P$. The intellectual challenge is to understand the lowest cost to ensure the optimal monotone error $k^*$ or an error higher than $k^*$ by a small multiplicative factor.

\boxminipg{0.95\linewidth}{
\begin{problem} \label{def:intro-prob1}
    {\em (\textsc{Monotone Classification with Relative Precision Guarantees})} Given a real value $\eps \ge 0$, find a monotone classifier whose error on $P$ is at most $(1+\eps) k^*$. The efficiency of an algorithm is measured by the number of elements probed.
\end{problem}
}

If $k^* = 0$, we say that $P$ is {\em monotone} and Problem 1 is {\em realizable}. Otherwise, $P$ is {\em non-monotone} and the problem is {\em non-realizable}.

\vgap

Every deterministic algorithm $\A_\det$ for Problem 1 can be modeled as a binary decision tree $\T$ that is determined by the coordinates of the elements in $P$. Each internal node of $\T$ is associated with an element in $P$, while each leaf of $\T$ is associated with a monotone classifier. The execution of $\A_\det$ descends a single root-to-leaf path in $\T$. When $\A_\det$ is at an internal node, it probes the element $p \in P$ associated with that node and branches left if $\lab(p) = -1$ or right if $\lab(p) = 1$. When the algorithm reaches a leaf of $\T$, it outputs the monotone classifier associated with the leaf.

\vgap

Randomized algorithms have access to an infinite bit string where each bit is independently 0 or 1 with probability $1/2$. A randomized algorithm $\A_\ran$ can be modeled as a function that maps the bit string to a deterministic algorithm (that is, when all the random bits are fixed, $\A_\ran$ specializes into a deterministic algorithm). Suppose that $h$ is the classifier output by $\A_\ran$ when executed on $P$, and $X$ is the number of probes \rev{performed by $\A_\ran$ on $P$}. Both $X$ and $h$ are random variables. The {\em expected cost} of $\A_\ran$ is defined as $\expt[X]$, while the {\em expected error} of $\A_\ran$ is defined as $\expt[\err_P(h)]$. We say that $\A_\ran$ guarantees an error $k$ {\em with high probability} (w.h.p.) if $\Pr[\err_P(h) \le k] \ge 1 - 1/n^c$ where $c$ can be an arbitrarily large constant set before running the algorithm.

\vgap

Although CPU time is not a main concern in this article, all proposed algorithms can be implemented to run in time \rev{polynomial in $n$ and $d$}, as will be duly noted in the technical development.

\subsection{Practical Motivations} \label{sec:intro:motivation}

An important application of monotone classification is ``entity matching''. Given two sets of entities $E_1$ and $E_2$, the goal is to decide, for each pair $(e_1, e_2) \in E_1 \times E_2$, whether $e_1$ and $e_2$ represent the same entity; if so, they are said to form a ``match''. For example, $E_1$ (resp., $E_2$) may be a set of advertisements placed on Amazon (resp., eBay). Each advertisement includes attributes like \ttt{prod-name}, \ttt{prod-description}, \ttt{year}, \ttt{price}, and so on. The goal is to identify the pairs $(e_1, e_2) \in E_1 \times E_2$ where the advertisements $e_1$ and $e_2$ describe the same product.

\vgap

What makes the problem challenging is that decisions cannot rely on comparing attribute values, because even a pair of matching $e_1$ and $e_2$ may differ in attributes. This is evident with attributes like \ttt{prod-description} and \ttt{price} since $e_1$ and $e_2$ might describe or price the same product differently. In fact, $e_1$ and $e_2$ may not even agree on ``presumably standard'' attributes like \ttt{prod-name} (e.g., $e_1.\ttt{prod-name}$ = ``MS Word'' vs.\ $e_2.\ttt{prod-name}$ = ``Microsoft Word Processor''). Nevertheless, it would be reasonable to expect $e_1.\ttt{year} = e_2.\ttt{year}$ because advertisements are required to be accurate in this respect. To attain full precision in entity matching, human experts must manually inspect each pair $(e_1, e_2) \in E_1 \times E_2$, which is expensive due to the intensive labor involved. Therefore, it is crucial to develop an algorithm that can minimize human effort by automatically rendering verdicts on most pairs, even if it involves a small margin of error.

\vgap

Toward the above purpose, a dominant methodology behind the existing approaches (e.g., \cite{agk10, cvw15, sb02, bipr13, cik16, epp+17, gdd+14, ktr10, tr07,vpss20}) is to transform the task into a multidimensional classification problem with the following preprocessing.

\vgap

\begin{enumerate}
    \item First, shrink the set of all possible pairs to a subset $S \subseteq E_1 \times E_2$, by eliminating pairs that clearly are not matches. This is known as {\em blocking}, which is carried out based on application-dependent heuristics. This step is optional; if skipped, then $S = E_1 \times E_2$. In the Amazon-eBay example, $S$ may involve only those advertisement pairs $(e_1, e_2)$ with $e_1.\ttt{year} = e_2.\ttt{year}$.

    \vgap

    \item For each entity pair $(e_1, e_2) \in S$, create a multidimensional point $p_{e_1, e_2}$ using several --- say $d$ --- similarity functions $\siml_1, \siml_2, ..., \siml_d$, each of which is evaluated on a certain attribute and produces a numeric ``feature''. The $i$-th coordinate of $p_{e_1,e_2}$ equals $\siml_i(e_1,e_2)$: a greater value indicates higher similarity between $e_1$ and $e_2$ under the $i$-th feature. This creates a $d$-dimensional point set $P = \{p_{e_1,e_2} \mid (e_1, e_2) \in S\}$. In our example, from a numerical attribute such as \ttt{price}, one may extract a similarity feature $-|e_1.\ttt{price} - e_2.\ttt{price}|$, where the negation is needed to ensure ``the larger the more similar''. From a text attribute (such as \ttt{prod-name} and \ttt{prod-description}) one may extract a similarity feature by evaluating the relevance between the corresponding texts of $e_1$ and $e_2$ using an appropriate metric (e.g., edit distance, jaccard-distance, cosine similarity, etc.). Multiple features may even be derived on the same attribute; e.g., one can extract two similarity features by computing the edit-distance and jaccard-distance of $e_1.\ttt{prod-name}$ and $e_2.\ttt{prod-name}$ separately.

    \vgap

    \item Every point $p_{e_1,e_2} \in P$ inherently carries a label, which is 1 if $(e_1,e_2)$ is a match, or $-1$ otherwise. The original entity matching task on $E_1$ and $E_2$ is now converted to inferring the labels of the points in $P$. Human inspection is the ultimate resort for determining the label of each $p_{e_1,e_2}$ with guaranteed correctness.
\end{enumerate}

\vgap

Treating $P$ as the input for monotone classification, we can employ an effective algorithm $\A$ designed for Problem~\ref{def:intro-prob1} to significantly reduce human labor. Specifically, the human plays the role of oracle: when given a point $p_{e_1,e_2}$, the human ``reveals'' the label of $p_{e_1,e_2}$ by manually checking whether $e_1$ and $e_2$ are about the same entity. After a  number of probes (to the human oracle), the algorithm $\A$ outputs a monotone classifier $h \in \mono$, which is then used to infer the labels of the un-probed points in $P$. Such a classifier is also suitable for performing matching on entities received in the future (assuming that $E_1$ and $E_2$ are representative of the underlying data distributions). Demanding monotonicity is important for {\em explainable learning} because it avoids the odd situation of classifying $(e_1,e_2)$ as a non-match but $(e_1',e_2')$ as a match when $p_{e_1,e_2} \dom p_{e_1',e_2'}$. Indeed, this oddity is difficult to explain because the former pair is at least as similar as the latter on every feature.

\begin{figure}
    \centering
    \includegraphics[height=45mm]{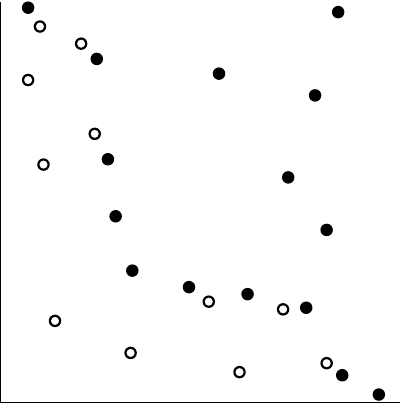}
    \figcapup
    \caption{\revv{A monotone input set that cannot be separated by linear classifiers}}
    \label{fig:intro-ds2}
    \figcapdown
\end{figure}

\revv{
We remark that, in entity-matching applications such as the one described above, it would also be reasonable to consider a restrictive subset of monotone classifiers. For example, one may resort to {\em linear classifiers}, each of which is parameterized by a $(d+1)$-dimensional vector $\bm{w}$ and maps a point $p \in \real^d$ to label 1 if $\sum_{i=1}^d \bm{w}[i] \cdot p[i] \ge \bm{w}[d+1]$ (note: $\bm{w}[i]$ denotes the $i$-th component of $\bm{w}$ for $i \in [d+1]$) or label $-1$ if $\sum_{i=1}^d \bm{w}[i] \cdot p[i] < \bm{w}[d+1]$. Equivalently, the classifier maps $p$ to 1 if it falls on or on the ``positive side'' of a $d$-dimensional plane or to $-1$ if it falls on the plane's ``negative'' side. Problem 1 can be re-defined with respect to the set of linear classifiers that obey monotonicity (as discussed earlier, monotonicity ensures explainable learning), by redefining $k^*$ in \eqref{eqn:k-star} as the smallest error of all such classifiers. In this work, we allow {\em arbitrary} monotone classifiers to maximize the ``classification power''. For instance, the set $P$ in Figure~\ref{fig:intro-ds2} --- where the black and white points have labels 1 and $-1$, respectively --- is monotone, whereas no linear classifier can separate the points of the two labels.
}


\subsection{Related Work} \label{sec:intro:prev}


\noindent {\bf Active Classification.} Classification is a fundamental topic in machine learning. In the standard setting, we consider a (possibly infinite) set $\P$ of points in $\real^d$ and an unknown distribution $\D$ over $\P \times \set{-1, 1}$. Given a sample $(p, l)$ drawn from $\D$, we refer to $p$ and $l$ as the sample's {\em point} and {\em label}, respectively. A {\em classifier} is a function $h: \P \rightarrow \set{-1, 1}$, whose {{\em error rate}} with respect to $\D$ is calculated as
\myeqn{
    \errate_\D (h) &=& \Pr_{(p, l) \sim \D} [h(p) \ne l]. \nn 
}
Let $\bbH$ denote the set of classifiers under consideration, for which the optimal error rate is
\myeqn{
    \nu = \inf_{h \in \bbH} \errate_\D(h). \label{eqn:nu}
}
The goal is to ensure the following {{\em probabilistically approximately correct}} (PAC) guarantee:

\begin{center}
    \begin{minipage}{0.9\linewidth}
        With probability at least $1 - \delta$, find a classifier $h \in \bbH$ such that $\errate_\D(h) \le \nu + \xi$ where $\xi$ and $\delta$ are problem parameters satisfying $0 < \xi < 1$ and $0 < \delta < 1$.
    \end{minipage}
\end{center}

An algorithm $\A$ is permitted to sample from $\D$ repeatedly. In {\em passive classification}, each sample $(p, l)$ drawn from $\D$ reveals both $p$ and $l$ directly. The performance of $\A$ is measured by its {\em sample cost}, defined as the number of samples taken. In many practical applications, producing the point field of a sample requires negligible cost, but determining its label is expensive. This motivates {\em active classification}, which aims to achieve the PAC guarantee without acquiring the labels of all samples. In this setup, when an algorithm $\A$ draws a sample $(p, l)$ from $\D$, only the point $p$ is shown to $\A$, but the label $l$ is hidden. The algorithm can request $l$ from an {\em oracle}, thereby doing a {\em probe}. If $\A$ considers the knowledge of $l$ unnecessary, it may skip the probe. The performance of $\A$ is now measured by its {{\em label cost}}, defined as the number of probes performed.

\vgap

\rev{Problem 1 is relevant to {\em \rev{non-realizable} active classification} where $\nu > 0$, i.e., even the best classifier in $\bbH$ has a positive, unknown error rate. This topic has been extensively studied; we refer the reader to \cite{bbl09,bdl09,d05,dhm07,h07b,h14,h25,hy15,k06,s10,w11} for representative works that serve as good entry points into the literature. By combining \cite{bbl09,h07b,h14,hy15}, one can achieve the PAC guarantee with a label cost of
\myeqn{
    \tilde{O}\left(\min\set{\theta, \starnum, \chi} \cdot \lambda \cdot \fr{\nu^2}{\xi^2} \right) \label{eqn:related-agnostic-prev1}
}
where $\lambda, \theta, \starnum$, and $\chi$ are the {\em VC-dimension}, {\em disagreement coefficient}, {\em star number}, and {\em teaching dimension} of $\bbH$ on $\P$, respectively, and $\tilde{O}(.)$ hides a polylogarithmic factor. The definitions of $\lambda, \theta, \starnum$, and $\chi$ are given in Appendix~\ref{app:vc-dc-of-mono}; for now, it suffices to understand that they are all at least 1. It is worth mentioning that the recent work of \cite{h25} described a method that can return a classifier $h$ satisfying $\errate_\D(h) \le \nu + \xi$ with a label cost of
    $O(\fr{\nu^2}{\xi^2} \cdot ( \lambda + \log \fr{1}{\delta}  )) + \tO (\min \{\starnum, 1/\xi \} \cdot \lambda )$
except that the classifier $h$ is not necessarily monotone.}

\vgap

\rev{
In the context of Problem 1, \uline{assuming that the optimal monotone error $k^*$ is known}, we can utilize the above results to settle the problem by
\myitems{
    \item setting $\bbH$ to $\mono$;
    \item setting $\P$ to the input $P$ of Problem 1;
    \item defining $\D$ as the uniform distribution over $\set{(p, \lab(p)) \mid p \in P}$.
}
Consequently, every monotone classifier $h$ satisfies $\errate_\D(h) = \err_P(h)/n$ where $n = |P|$ and $\err_P(h)$ is given in \eqref{eqn:err-classifier}. In other words:
\myeqn{
    \nu = k^* / n. \label{eqn:related-set-nu}
}
Thus, by choosing
\myeqn{
    \xi = \eps \cdot k^* / n \label{eqn:related-set-xi}
}
and applying the result in \eqref{eqn:related-agnostic-prev1}, we can find a $(1+\eps)$-approximate monotone classifier w.h.p.\ with a cost bounded by
\myeqn{
    \tilde{O}\left(\min\set{\theta, \starnum, \chi} \cdot \lambda \cdot \fr{1}{\eps^2} \right). \label{eqn:related-prev1}
}


\rev{
When $k^*$ is unknown, however, one can no longer set $\xi$ ``correctly'' as in \eqref{eqn:related-set-xi}. While the algorithms of \cite{bbl09,h07b,h14,h25,hy15} require no knowledge of $\nu$ (and, hence, $k^*$) to run, they all demand a concrete $\xi$. One potential remedy would be to start with a large $\xi$ (e.g., $0.5$) and then iteratively reduce it until $\xi$ reaches the value in \eqref{eqn:related-set-xi}. However, it is difficult (if not impossible) to apply such a remedy to the existing algorithms. A source of difficulty is that those algorithms work by bounding the {\em discrepancy} between an optimal classifier $h^*$ and the returned classifier $h$, ensuring that the discrepancy can cause an ``excess'' error rate at most $\xi$. They cannot afford to infer the error rate caused by the points on which $h^*$ and $h$ assign the same labels --- doing so would incur the same cost as passive learning. As a result, those algorithms do not have a reliable estimate about $\nu$, making it difficult for them to decide whether $\xi$ has come close to \eqref{eqn:related-set-xi}. In fact, to genuinely settle Problem 1 (with no assumptions), an active learning algorithm would need to return a classifier with an error rate at most $\nu (1+\eps)$ for a given value of $\eps$. We are not aware of any such algorithms.
}

\begin{figure}
    \centering
    \begin{tabular}{cc}
        \includegraphics[height=20mm]{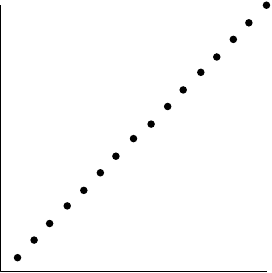} &
        \includegraphics[height=20mm]{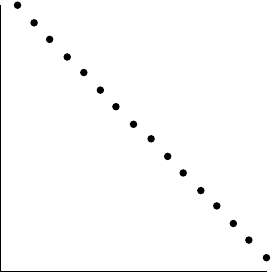} \\
        (a) A point set of width 1 & (b) A point set of width $n$
    \end{tabular}
    \figcapup
    \caption{Illustration of the dominance width}
    \label{fig:intro:width}
    \figcapdown
\end{figure}

\vgap

\rev{{\em Hypercube learning} is a special case of monotone classification (i.e., the classification problem with $\bbH = \mono$) that has received particular attention.} Here, $\P = \set{0, 1}^d$, every point $p \in \P$ carries a label $\lab(p)$, and $\D$ is the uniform distribution over $\{(p, \lab(p) \mid p \in P\}$. Improving over \cite{bt96,lrv22}, Lange and Vasilyan \cite{lv23} achieved the PAC guarantee with a sample cost \rev{(hence, also a label cost)} of $2^{\tilde{O}(\sqrt{d} / \xi)}$, where $\tilde{O}(.)$ hides a logarithmic factor.

\extraspacing {\bf Monotonicity Testing.} Given a multiset $P$ of $n$ labeled points as defined in Problem 1, {{\em monotonicity testing}} \rev{ --- a special instance of the general framework of {\em property testing} ---} is the problem of deciding whether $P$ is monotone (i.e., having optimal monotone error $k^* = 0$) or far from being so. Formally, given an input parameter $\xi > 0$, the output must always be ``yes'' if $P$ is monotone. If $k^* \ge \xi n$, the output should be ``no'' with probability at least $2/3$. In the scenario where $0 < k^* < \xi n$, the output can be either way. In the beginning, all the point labels are hidden. An algorithm $\A$ interacts with an oracle in the same manner as in Problem 1; its performance is measured by the number of probes carried out. Fischer et al.\ \cite{fln+02} gave an algorithm that solves the problem with $O(\sqrt{n/\xi})$ probes. The problem has also been explored in other settings less relevant to this article; the interested readers may refer to \cite{cs13,cs14,cdh+25,ggl+00} and the references therein.

\subsection{Our Results} \label{sec:intro:ours}

For $\eps = 0$, Problem~\ref{def:intro-prob1} aims to find an optimal monotone classifier. We prove that any algorithm achieving the purpose with a probability \rev{greater than 2/3} needs to probe $\Omega(n)$ points in expectation (Theorem~\ref{thm:lb-eps0}). The lower bound holds even when the dimensionality $d$ is 1 and the optimal monotone error $k^*$ is known to the algorithm.

\vgap

The above hardness result justifies an investigation of Problem 1 under $\eps > 0$. In this regime, we show that the probing complexity is determined by the {\em width} $w$ of the input $P$, \rev{formally defined} as:
\myeqn{
    w &=& \text{the size of the largest $S \subseteq P$ such that $\nexists$ distinct $p, q \in S$ satisfying $p \domeq q$}. \label{eqn:intro:w}
}
Any one-dimensional $P$ has $w = 1$. Once the dimensionality $d$ reaches 2, $w$ can be anywhere between $1$ and $n$; see Figure~\ref{fig:intro:width} for two extreme examples.

\begin{example}
    Regarding the input $P$ in Figure~\ref{fig:intro-ds}, the set $S = \{p_{10}, p_{11}, p_{12}, p_{16}, p_{13},$ $p_{14}\}$ satisfies the condition on the right hand side of \eqref{eqn:intro:w}. On the other hand, given any 7 points of $P$, we can always find two points where one point dominates the other. The dominance width $w$ of $P$ is 6. 
\end{example}

Equipped with $w$, we can \rev{expand our discussion on active learning in the context of Problem 1. As explained in Section~\ref{sec:intro:prev}, the existing algorithms are applicable only if $k^*$ is known. While this is an unreasonable  assumption, for the sake of a qualitative comparison, let us favor those algorithms by making $k^*$ available for free, thus enabling them to achieve a label cost bounded by \eqref{eqn:related-prev1}. In Appendix~\ref{app:vc-dc-of-mono}, we prove that $\lambda$ (the VC-dimension), $\theta$ (the disagreement coefficient), $\starnum$ (the star number), and $\chi$ (the teaching dimension) can all be $\Omega(w)$. As a result, the best complexity from \eqref{eqn:related-prev1} is
    $\tO (w^2 /\eps^2)
$
where $\tO(.)$ hides a polylogarithmic factor.
}

\begin{table}
    \begin{tabular}{c|c|c|c}
        $\bm{\eps}$ & {\bf probing cost} & {\bf refs} & {\bf remarks}
        \\
        \hline\hline
        any & $\tO(w^2 / \eps^2)$ \rev{if $k^*$ known} & \rev{\cite{bbl09,h07b,h14,hy15}} & w.h.p.\ error
        \\
        any &
        \rev{
        $\tO(
        \fr{w}{\eps^2} + \min \{
            w^2, \fr{n \cdot w}{\eps \cdot k^*}
        \}
        )$ if $k^*$ known}
        & \cite{h25} &
        w.h.p.\ error
        \\
        \hline\hline
        1 & $O(w \Log \fr{n}{w})$ expected & Thm.\ \ref{thm:rpe} & expected error
        \\
        any & $O(\fr{w}{\eps^2} \cdot \Log \fr{n}{w} \cdot \log n)$ & Thm.\ \ref{thm:coreset} + Sec.\ \ref{sec:intro:techniques} & w.h.p.\ error
        \\
        \hline
        0 & $\Omega(n)$ & Thm.\ \ref{thm:lb-eps0} & success probability $> 2/3$
        \\
        constant & $\Omega(w \Log \fr{n}{(k^*+1) w})$ expected & Thm.\ \ref{thm:lb-eps-const} & expected error \\
        any & $\Omega(w / \eps^2)$ expected & Thm.\ \ref{thm:lb-eps-any} & expected error
    \end{tabular}
    \caption{A summary of our and previous results on Problem 1}
    \label{tab:intro:results}
\end{table}

\vgap

This article shows that the ``true'' complexity of Problem 1 is \rev{on the order of $w/\eps^2$ up to polylogarithmic factors}. We achieve the purpose by establishing nearly-matching upper and lower bounds. Regarding upper bounds:
\myitems{
    \item We design an algorithm that guarantees an expected error at most $2 k^*$ with an expected cost of $O(w \Log \fr{n}{w})$ (Theorem~\ref{thm:rpe}). When $k^* = 0$ --- i.e., Problem 1 is realizable --- the algorithm always finds an optimal classifier.

    \item For any $\eps > 0$, we design an algorithm that guarantees an error of $(1+\eps) k^*$ w.h.p.\ at a cost of $O(\fr{w}{\eps^2} \cdot \Log \fr{n}{w} \cdot \log n)$ (Theorem~\ref{thm:coreset}).
}
\rev{Regarding} lower bounds:
\myitems{
    \item For any constant $c > 1$, we prove that any algorithm ensuring an expected error at most $c k^*$ must probe $\Omega(w \Log \fr{n}{(k^*+1) w})$ points in expectation (Theorem~\ref{thm:lb-eps-const}). Our first algorithm (with expected error at most $2k^*$) is thus asymptotically optimal when $k^* \le (n/w)^{1-\delta}$ where $\delta > 0$ is an arbitrarily small constant.

    \item For any $\eps > 0$, we prove that any algorithm ensuring an expected error at most $(1+\eps) k^*$ must probe $\Omega(w / \eps^2)$ points in expectation (Theorem~\ref{thm:lb-eps-any}). Hence, our second algorithm is asymptotically optimal up to a factor of $O(\Log \fr{n}{w} \cdot \log n)$.
}
Table~\ref{tab:intro:results} summarizes all the results mentioned earlier.

\vgap

As a side product, our findings imply a new result for monotonicity testing. Recall from Section~\ref{sec:intro:prev} that the input to monotonicity testing includes a multiset $P$ of $n$ labeled points and a real-valued parameter $\xi \in (0, 1)$. In Section~\ref{sec:rpe:testing}, we solve the problem with an expected  cost of $O(w \Log \fr{n}{w}) + 2 / \xi$, where $w$ is the width of $P$. \rev{Recall that the state of the art on this problem probes $O(\sqrt{n / \xi})$ points \cite{gglr98}. Our solution is expected to probe less points when $w$ is small.}

\vgap

\rev{
Preliminary versions of this work appeared in PODS'18 \cite{t18} and PODS'21 \cite{tw21}. In terms of technical content, the current article differs from those papers in four ways: (i) we present a new, simpler analysis of the \ttt{RPE} algorithm (see Section~\ref{sec:intro:techniques}); (ii) we establish a new coreset result (Theorem~\ref{thm:coreset}); (iii) we prove a lower bound for Problem 1 under arbitrary $\eps$ values; (iv) we strengthen the relevance to active learning, including the derivation of lower bounds for $\lambda$, $\theta$, $\starnum$, and $\chi$ in the setting of Problem 1.
}

\subsection{\rev{Overview of Our Algorithms}} \label{sec:intro:techniques}

\rev{This subsection highlights the algorithmic techniques developed in this work.}

\extraspacing {\bf \rev{A Randomized Algorithm.}} Our first algorithm --- named \ttt{RPE} (random probes with elimination) --- \rev{is rather simple}:

\mytab{
        \> {\bf algorithm \ttt{RPE}} $(P)$ \\
        \> 1. \> $Z = \emptyset$ \\
        \> 2. \> {\bf while} $P \neq \emptyset$ \\
        \> 3. \>\> probe an element $z \in P$ chosen uniformly at random and add $z$ to $Z$ \\
        \> 4. \>\> {\bf if} $\lab(z) = 1$ {\bf then} remove from $P$ every $p \in P$ satisfying $p \domeq z$ \\
        \> 5. \>\> {\bf else} remove from $P$ every $p \in P$ satisfying $z \domeq p$ \\
        \> \>\> /* note: $z$ is removed by Line 4 or 5 */ \\
        \> 6. \> {\bf return} $Z$
}

\noindent The algorithm outputs the set $Z$ of points probed. Given $Z$, we define the following classifier:
\myeqn{
    h_\rpe(p) &=& \left\{
    \begin{tabular}{ll}
        1 & if $\exists z \in Z$ such that $\lab(z) = 1$ and $p \domeq z$ \\
        $-1$ & otherwise
    \end{tabular}
    \right. \label{eqn:rpe-classifier}
}
The classifier must be monotone. Otherwise, there exist $p, q \in \real^d$ such that $p \dom q$ but $h_\rpe(p) = -1$ and $h_\rpe(q) = 1$. Let $z$ be an arbitrary element in $Z$ satisfying $\lab(z) = 1$ and $q \domeq z$ (such $z$ must exist by the definition in \eqref{eqn:rpe-classifier}). It follows that $p \dom z$, in which case $h_\rpe(p)$ should be 1 according to \eqref{eqn:rpe-classifier}, giving a contradiction.

\vgap

\ttt{RPE} can also be understood as the following labeling process. After learning the label of a random element $z \in P$, we decide to map $z$ to $\lab(z)$. If $\lab(z) = 1$, by monotonicity, we are obliged to map all elements $p \in P$ satisfying $p \domeq z$ to $1$; conversely, if $\lab(z) = -1$, we must map all elements $p \in P$ satisfying $z \domeq p$ to $-1$. The elements that have their labels thus mapped are removed from $P$. The process is then repeated on the remaining elements.

\begin{example}
    Let us illustrate \ttt{RPE} on the input $P$ in Figure~\ref{fig:intro-ds}. Assume that the algorithm randomly probes $p_1$ first. Acquiring $\lab(p_1) = 1$, it eliminates the entire $P$ except $p_6, p_7$, and $p_8$. Suppose that \ttt{RPE} then randomly probes $p_8$. Acquiring $\lab(p_8) = -1$, it removes all the remaining points in $P$. With $Z = \{p_1, p_8\}$, the classifier $h_\rpe$ in \eqref{eqn:rpe-classifier} maps all the points to label 1 except $p_6, p_7$, and $p_8$. Its error on $P$ is $\err_P(h_\rpe) = 5$.
\end{example}

\rev{It may be unclear why the classifier in \eqref{eqn:rpe-classifier} ensures an expected error at most $2k^*$ --- after all,} probing a ``wrong'' element of $P$ may immediately force $h_\rpe$ to misclassify a large number of elements (e.g., imagine what if $p_{15}$ is probed first in Figure~\ref{fig:intro-ds}). \rev{Furthermore, it may be unclear why \ttt{RPE} probes} $O(w \Log \fr{n}{w})$ elements in expectation, particularly since the algorithm never computes the value of $w$. \rev{We answer these questions in Section~\ref{sec:rpe}.}


\extraspacing {\bf Relative-Comparison Coresets.} \rev{The approximation ratio 2 is tight for \ttt{RPE} (this will be explained in Section~\ref{sec:rpe:error}). To achieve an approximation ratio of $1 + \eps$ for any $\eps > 0$, we aim to} produce a function $F: \mono \rightarrow \real$ having the property below.

\vgap

\minipg{0.9\linewidth}{
    {{\em The relative $\eps$-comparison property}}: $F(h) \le F(h')$ implies $\rev{\err_P(h) \le_{1+\eps} \err_P(h')}$ for any classifiers $h, h' \in \mono$.
}

\vgap

\noindent Given $F$, we can identify a monotone classifier $h^\newstar$ with the lowest $F(h^\newstar)$. This classifier satisfies $\err_P(h^\newstar) \le (1 + \eps) k^*$ (noticing that $F(h^\newstar) \le F(h^*)$ where $h^*$ is an optimal classifier on $P$, i.e, $\err_P(h^*) = k^*$) and thus can be returned as an output of Problem 1.

\vgap

A barrier in finding such a function $F$ is that we cannot hope to \rev{accurately} estimate the $\err_P(h)$ of every monotone classifier $h$. This is true even if the dimensionality is 1. To see why, consider the classifier $h^\pos$ that maps everything to 1. Its error is 0 if all elements of $P$ have label 1, or error 1 if all but one element in $P$ has label 1. Hence, estimating $\err_P(h^\pos)$ up to a relative ratio, say, $1/2$ requires identifying the only element in $P$ having label $-1$ or declaring the absence of such an element. It is not hard to prove that achieving the purpose with a constant probability demands $\Omega(n)$ probes in expectation.

\vgap

Instead, we produce $F$ by ensuring the following inequality on every $h \in \mono$:
\myeqn{
    \err_P(h) \cdot \left(1-\fr{\eps}{4}\right) + \Delta
    \le
    F(h)
    \le
    \err_P(h) \cdot \left(1+\fr{\eps}{4}\right) + \Delta
    \label{eqn:intro:F-property}
}
where $\Delta$ is an \uline{unknown} real value common to all $h$. If the exact value of $\Delta$ {\em were} available, then $F(h) - \Delta$ {\em would} serve as an estimate of $\err_P(h)$ with a relative ratio at most $\eps/4$, which would require $\Omega(n)$ probes as discussed. The key behind our \rev{solution} is to compute $F(h)$ {\em without} knowing $\Delta$, as long as the {\em existence} of $\Delta$ can be assured.

\vgap

The relative $\eps$-comparison property is a corollary of \eqref{eqn:intro:F-property} because
\myeqn{
    \err_P(h)
    &\le&
    \fr{1}{1-\eps/4} \cdot (F(h) - \Delta) \nn \\
    \textrm{(by $F(h) \le F(h')$)}
    &\le&
    \fr{1}{1-\eps/4} \cdot (F(h') - \Delta) \nn \\
    \textrm{(applying \eqref{eqn:intro:F-property})}
    &\le&
    \fr{1+\eps/4}{1-\eps/4} \cdot err_P(h') \nn \\
    &\le&
    (1+\eps) \cdot err_P(h') \nn
}
where the last inequality used the fact that $\fr{1+\eps/4}{1-\eps/4} \le 1+\eps$ for all $\eps \in (0, 1]$. The existence of $\Delta$ already permits the derivation to proceed --- its concrete value is unnecessary.

\vgap

\rev{We derive a function $F$ meeting the condition in \eqref{eqn:intro:F-property} from a {\em coreset} of $P$}, which is a subset $Z \subseteq P$ where each element $p \in Z$
\myitems{
    \item has its label revealed, and
    \item is associated with a positive real value $\weight(p)$, called the {\em weight} of $p$.
}
Given a classifier $h \in \mono$, define its {{\em weighted error}} on $Z$ as:
\myeqn{
    \werr_Z(h)
    &=&
    \sum_{p \in Z} \weight(p) \cdot \mathbbm{1}_{h(p) \neq \lab(p)}.
    \label{eqn:intro:weighted-err}
}
With $O(\fr{w}{\eps^2} \Log \fr{n}{w} \cdot \log n)$ probes, we can obtain w.h.p.\ a coreset $Z$ of size $O(\fr{w}{\eps^2} \Log \fr{n}{w} \cdot \log n)$, such that every $h \in \mono$ satisfies
\myeqn{
    \err_P(h) \cdot \left(1-\fr{\eps}{4}\right) + \Delta
    \le
    \werr_Z(h)
    \le
    \err_P(h) \cdot \left(1+\fr{\eps}{4}\right) + \Delta
    \label{eqn:intro:F-property-thru-coreset}
}
for some unknown $\Delta$. The function $\werr_Z$ serves as the desired function $F$. We refer to $Z$ as a {{\em relative-comparison coreset}} because its purpose is to enable the relative $\eps$-comparison property.

\vgap


\revv{There is a large body of work on coresets --- see  \cite{ap24,blk18,bfl+21,c09,ddh+09,fl11,fss20,hk07,hs11,p16b} and the references therein --- which generally refers to a small set of elements that approximately preserves a specified property of the original input set. Our target of preservation --- namely, the inequalities in \eqref{eqn:intro:F-property} --- is closely related to the coreset formulation in \cite{fss20} (see in particular Definition 13 therein), which likewise uses an additive $\Delta$ to support relative comparisons, albeit in a different setting. As will be clear in Section~\ref{sec:coreset}, a key challenge in our analysis is to argue for the {\em existence} of a working $\Delta$ in \eqref{eqn:intro:F-property} without explicitly computing it --- as noted before, deriving $\Delta$ precisely demands $\Omega(n)$ probes. The idea of using an ``unknown parameter'' such as $\Delta$ for coreset construction is not new; e.g., it is implicit in the sampling technique of \cite{mmr19} for clustering polygonal curves. Our contribution is to make the idea work in the context of finding a good monotone classifier.}

\vgap

\revv{We close the section with a remark on the relevance of our work to coresets in machine learning that are designed for approximate regression within a relative factor (see  \cite{ap24,mmr21,mssw18,spm+20,tjf22} and the references therein). As discussed in Section~\ref{sec:intro:motivation}, Problem 1 can be re-defined as a problem over any restrictive subset of monotone classifiers, e.g., linear classifiers that obey monotonicity. Because, as explained before, the function $F$ we construct guarantees the relative $\eps$-comparison property for all monotone classifiers, it (trivially) offers such a guarantee for all linear monotone classifiers. Therefore, our method also solves Problem 1 with respect to linear monotone classifiers at the same probing cost (more generally, the same holds for any subset of monotone classifiers). Although linear classifiers are well-studied, we are not aware of any coreset result that can match ours for Problem 1. Perhaps the most related result in this regard is the recent work of \cite{ap24}, which proves the existence of  small coresets for linear classifiers under logistic regression. Recall that the imprecision measure in Problem 1 (see \eqref{eqn:err-classifier}) is the standard ``0-1 error'', for which the logistic loss (i.e., the loss function to be minimized in logistic regression) is a widely used surrogate in practice. In theory, however, a relative-$\eps$ approximation of logistic loss does not imply a relative-$\eps$ approximation of the 0-1 error. Thus, the guarantee of \cite{ap24} does not imply the guarantee required by Problem 1, even for linear monotone classifiers. Similar observations apply to coresets designed to approximate other surrogate objectives.}

\section{Random Probes with Elimination} \label{sec:rpe}

This section proves the following theorem on the \ttt{RPE} algorithm in Section~\ref{sec:intro:techniques}:

\begin{theorem} \label{thm:rpe}
    For Problem 1, \ttt{RPE} probes $O(w \Log \fr{n}{w})$ elements in expectation, and the classifier $h_\rpe$ in \eqref{eqn:rpe-classifier} has an expected error at most $2 k^*$, where $n$ is the size of the input $P$, $w$ is its width, and $k^*$ is its optimal monotone error.
\end{theorem}

Our proof is divided into two parts: Section~\ref{sec:rpe:error} analyzes the expected error, and Section~\ref{sec:rpe:cost} examines the expected cost. The following proposition states a useful property of the classifier $h_\rpe$, establishing symmetry between labels $-1$ and $1$.

\begin{proposition} \label{prop:rpe:basic}
    For any $p \in P$, $h_\rpe(p)$ $=$ $-1$ if and only if $\exists z \in Z$ satisfying $\lab(z) = -1$ and $z \domeq p$, where $Z$ is the set of elements probed by \ttt{RPE}.
\end{proposition}

\begin{proof}
    We first show that $Z$ is monotone, namely:
    \myitems{
        \item $Z$ contains no two distinct elements $p$ and $q$ with $p = q$;
        \item for any $p, q \in Z$: $p \dom q \Rightarrow \lab(p) \ge \lab(q)$.
    }
    The first bullet holds because once an element $z \in P$ is probed at Line 3 of the pseudocode in Section~\ref{sec:intro:techniques},  all the elements of $P$ at the same location as $z$ are removed from $P$ at Line 4 or 5.   Regarding the second bullet, assume that there exist $p, q \in Z$ such that $p \dom q$, $\lab(p) = -1$, and $\lab(q) = 1$. Which element was probed earlier by \ttt{RPE}? If it was $p$, then $q$ should have been removed from $P$ at Line 5 after the probing of $p$ at Line 3, contradicting $q \in Z$. Likewise, probing $q$ first would contradict $p \in Z$.

    \vgap

    \rev{Next, we} prove that, for any $p \in P$, the condition $h_\rpe(p) = -1$ holds if and only if $\exists z \in Z$ satisfying $\lab(z) = -1$ and $z \domeq p$. Let us first consider the ``only-if direction'' (i.e., $\Rightarrow$).  Our argument proceeds differently in two cases.
    \myitems{
        \item Case $p \in Z$: This implies $\lab(p) = -1$ (indeed, if $\lab(p) = 1$, then $h_\rpe(p) = 1$ because we can $z = P$ in \eqref{eqn:rpe-classifier}). We can set $z = p$ to fulfill the only-if statement.

        \vgap

        \item Case $p \notin Z$: From $h_\rpe(p) = -1$, we can assert by \eqref{eqn:rpe-classifier} that no element $z' \in Z$ satisfies $\lab(z') = 1$ and $p \domeq z'$. Thus, \ttt{RPE} must have removed $p$ after probing an element $z$ satisfying $\lab(z) = -1$ and $z \domeq p$. This $z$ fulfills the only-if statement.
    }
    Finally, we consider the ``if direction'' (i.e., $\Leftarrow$).
    The designated element $z$ rules out the existence of any element $z' \in Z$ satisfying $\lab(z') = 1$ and $p \domeq z'$; otherwise, $z \domeq z'$ and the labels of $z$ and $z'$ suggest that $Z$ is not monotone. Hence, $h_\rpe(p) = -1$ by \eqref{eqn:rpe-classifier}.
\end{proof}

\subsection{The Expected Error of \ttt{RPE}} \label{sec:rpe:error}

Fix an arbitrary classifier $h \in \mono$. An element $p \in P$ is said to be {\em $h$-good} if $h(p) = \lab(p)$ or {\em $h$-bad} otherwise. We  prove in this subsection:

\begin{lemma} \label{lmm:rpe:h-good}
    The number of $h$-good elements misclassified by the classifier $h_\rpe$ in \eqref{eqn:rpe-classifier} is at most $\err_P(h)$ in expectation.
\end{lemma}

The lemma implies $\expt[\err_P(h_\rpe)] \le 2k^*$. To understand why, set $h$ to an optimal monotone classifier $h^*$ on $P$. By Lemma~\ref{lmm:rpe:h-good}, the number of $h^*$-good elements misclassified by $h_\rpe$ is at most $\err_P(h^*) = k^*$ in expectation. Because exactly $k^*$ elements of $P$ are $h^*$-bad, the total number of elements misclassified by $h_\rpe$ is at most $2k^*$ in expectation.

\vgap

Our proof of Lemma~\ref{lmm:rpe:h-good} works by induction on the size $n$ of $P$. When $n = 1$, the classifier $h_\rpe$ has error 0 on $P$, and the claim holds. Next, assuming that the claim holds for $n \le m - 1$ (where $m \ge 2$), we establish its correctness for $n = m$. Define
    \myeqn{
        X &=& \text{the number of $h$-good elements in $P$ misclassified by $h_\rpe$}, \label{eqn:rpe:X}
    }
    Our goal is to show that $\expt[X] \le \err_P(h)$.

    \vgap

    For each $h$-bad element $p$, we define its {\em influence set} $I_\bad(p)$ as follows:
    \myitems{
        \item If $\lab(p) = -1$, then $I_\bad(p)$ consists of all the $h$-good elements $q \in P$ satisfying $p \domeq q$ and $\lab(q) = 1$;
        \item If $\lab(p) = 1$, then $I_\bad(p)$ consists of all the $h$-good elements $q \in P$ satisfying $q \domeq p$ and $\lab(q) = -1$.
    }
    \rev{Intuitively, $I_\bad(p)$ includes all the $h$-good elements $q$ of $P$ such that once \ttt{RPE} probes $p$, it will mis-classify $q$.} For each $h$-good element $q \in P$, define its {\em influence set} $I_\good(q)$ as follows:
    \myitems{
        \item If $\lab(q) = -1$, then $I_\good(q)$ consists of all the $h$-bad elements $p \in P$ satisfying $q \domeq p$.

        \item If $\lab(q) = 1$, then $I_\good(q)$ consists of all the $h$-bad elements $p \in P$ satisfying $p \domeq q$.
    }
    \rev{Intuitively, $I_\good(q)$ includes all the $h$-bad elements $p$ of $P$ such that once \ttt{RPE} probes $q$, it will mis-classify $p$.}

\begin{lemma} \label{lmm:rpe:inf-set}
    \rev{For any $h$-bad $p \in P$ and any $h$-good $q \in P$, we have $ q \in I_\bad(p) \Leftrightarrow p \in I_\good(q)$.}
\end{lemma}

\begin{proof}
    \rev{First, we prove $q \in I_\bad(p) \Rightarrow p \in I_\good(q)$. By symmetry, we discuss only the case $\lab(p) = -1$. Because $q \in I_\bad(p)$, the definition of $I_\bad(p)$ tells us $p \domeq q$ and $\lab(q) = 1$. It thus follows that $p \in I_\good(q)$.}

    \vgap

    \rev{Next, we prove $p \in I_\good(q) \Rightarrow q \in I_\bad(p)$. By symmetry, we discuss only the case $\lab(q) = -1$. Because $p \in I_\good(q)$, the definition of $I_\good(q)$ tells us $q \domeq p$. Because $q$ is $h$-good, we know $h(q) = \lab(q) = -1$, which by monotonicity implies $h(p) = -1$. As $p$ is $h$-bad, we have $\lab(p) = 1$. It thus follows that $q \in I_\bad(p)$.}
\end{proof}

Lemma~\ref{lmm:rpe:inf-set} indicates
    \myeqn{
        \sum_{\text{$h$-good $q \in P$}} |I_\good(q)| &=& \sum_{\text{$h$-bad $p \in P$}} |I_\bad(p)|.
        \label{eqn:rpe:good-bad-equal}
    }

    \begin{example}
        Let $P$ be the set of points in Figure~\ref{fig:intro-ds}, and $h$ be the classifier that maps all the black points to 1 except $p_1$ and all the white points to $-1$ except $p_{11}$ and $p_{15}$. Thus, $p_1$, $p_{11}$, and $p_{15}$ are $h$-bad, while the other points of $P$ are $h$-good. The following are some representative influence sets.
        \myeqn{
            I_\bad(p_{15}) &=& \set{p_4, p_9, p_{10}, p_{12}, p_{13}, p_{14}, p_{16}} \nn \\
            I_\bad(p_1) &=& \set{p_2, p_3, p_5} \nn \\
            I_\good(p_3) &=& \set{p_1} \nn \\
            I_\good(p_9) &=& \set{p_{11}, p_{15}}. \nn
        }
    \end{example}

    \vgap

    \ttt{RPE} is an iterative algorithm. We refer to Lines 3-5 \rev{of its pseudocode} as an {\em iteration}. Let $z$ be the point probed in the first iteration (at Line 3). The revelation of $\lab(z)$ instructs the algorithm to remove $z$ and possibly some other elements from $P$ (at Line 4 or 5). Define
    \myeqn{
        P_z &=& \text{the set of remaining elements at the end of the first iteration.} \label{eqn:rpe:P_z}
    }
    The next proposition relates the error of $h$ on $P_z$ to the error of $h$ on $P$.
    \begin{proposition} \label{prop:rpe:err-Pz}
        \myeqn{
        \err_{P_z}(h) &\le& \left\{
        \begin{tabular}{ll}
            $\err_P(h) - 1$ & if $z$ is $h$-bad \\
            $\err_P(h) - |I_\good(z)|$ & if $z$ is $h$-good
        \end{tabular}
        \right. \nn
    }
    \end{proposition}

    \begin{proof}
        If $z$ is $h$-bad, the inequality $\err_{P_z}(h) \le \err_P(h) - 1$ follows immediately from the fact that $P_z$ has lost at least one $h$-bad element (i.e., $z$).

        \vgap

        Consider that $z$ is $h$-good. Assume $\lab(z) = 1$. In the first iteration, an element $p \in P$ is removed by Line 4 if and only if $p \domeq z$. By definition of $I_\good(z)$, an element $p$ removed by Line 4 is $h$-bad if and only if $p \in I_\good(z)$. Hence, $P_z$ loses exactly $|I_\good(z)|$ $h$-bad elements compared to $P$, giving $\err_{P_z}(h) = \err_P(h) - |I_\good(z)|$. A symmetric argument applies to the case where $\lab(z) = -1$.
    \end{proof}

    Define
    \myeqn{
        Y_z &=& \text{the number of $h$-good elements in $P_z$ misclassified by $h_\rpe$} \nn
    }
    Under the event that the first element probed is $z$, we have
    \myeqn{
        X &=& Y_z \, + \, \text{the number of $h$-good elements in $P \setm P_z$ misclassified by $h_\rpe$} \label{eqn:rpe:X-Y}
    }
    where $X$ is defined in \eqref{eqn:rpe:X}.

    \begin{proposition} \label{prop:rpe:err-P-Pz}
        The number of $h$-good elements in $P \setm P_z$ misclassified by $h_\rpe$ is
        \myitems{
            \item $|I_\bad(z)|$ if $z$ is $h$-bad;
            \item 0 if $z$ is $h$-good.
        }
    \end{proposition}

    \begin{proof}
        Consider the case $\lab(z) = -1$. The set $P \setm P_z$ consists of every element $p \in P$ satisfying $z \domeq p$. By Proposition~\ref{prop:rpe:basic}, the classifier $h_\rpe$ maps every such an element to $-1$. Hence, the number of $h$-good points in $P \setm P_z$ misclassified by $h_\rpe$ equals the number --- let it be $x$ --- of $h$-good points in $P \setm P_z$ whose labels are $1$. If $z$ is $h$-bad, then $x$ is exactly $|I_\bad(z)|$ by definition of $I_\bad(z)$. Suppose that $z$ is $h$-good. By monotonicity of $h$, we have $h(p) = -1$ for all $p \domeq z$ (recall that $\lab(z) = -1$). If $p$ is $h$-good, then $\lab(p) = h(p) = -1$. \rev{The number $x$} is thus 0 if $z$ is $h$-good.

        \vgap

        A symmetric argument applies to the case $\lab(z) = 1$.
    \end{proof}

    Combining \eqref{eqn:rpe:X-Y} and Proposition~\ref{prop:rpe:err-P-Pz} yields
    \myeqn{
        X &=& \left\{
        \begin{tabular}{ll}
            $Y_z + |I_\bad(z)|$ & if $z$ is $h$-bad \\
            $Y_z$ & if $z$ is $h$-good
        \end{tabular}
        \right. \label{eqn:rpe:X-Y-2}
    }

    The subsequent execution of \ttt{RPE} on $P_z$ can be regarded as invoking \ttt{RPE} directly on $P_z$. As $P_z$ has at least one less element than $P$ (because $z$ is removed), the inductive assumption tells us:
    \myeqn{
        \expt[Y_z] &\le& \err_{P_z}(h). \label{eqn:rpe:expt-Yz}
    }
    We can now derive
    \myeqn{
        \expt[X] &=& \sum_{z \in P} \expt[X \mid \text{$z$ is probed first}] \cdot \Pr[\text{$z$ is probed first}] \nn \\
        &=& \fr{1}{m} \sum_{z \in P} \expt[X \mid \text{$z$ is probed first}] \nn \\
        &=& \fr{1}{m} \Big( \sum_{\text{$h$-good $z \in P$}} \expt[X \mid \text{$z$ is probed first}] + \sum_{\text{$h$-bad $z \in P$}} \expt[X \mid \text{$z$ is probed first}]
        \Big) \nn \\
        \explain{by \eqref{eqn:rpe:X-Y-2}}
        &=& \Big(\fr{1}{m} \sum_{\text{$h$-good $z \in P$}} \expt[Y_z] \Big) + \Big( \fr{1}{m} \sum_{\text{$h$-bad $z \in P$}} \expt[Y_z] + |I_\bad(z)|
        \Big) \nn \\
        \explain{by \eqref{eqn:rpe:expt-Yz}}
        &\le&
        \Big(\fr{1}{m} \sum_{\text{$h$-good $z \in P$}} err_{P_z}(h) \Big) + \Big(\fr{1}{m} \sum_{\text{$h$-bad $z \in P$}} \err_{P_z}(h) + |I_\bad(z)|
        \Big) \nn \\
        \explain{Proposition~\ref{prop:rpe:err-Pz}}
        &\le&
        \Big(\fr{1}{m} \sum_{\text{$h$-good $z \in P$}} err_P(h) - |I_\good(z)| \Big) + \Big(\fr{1}{m} \sum_{\text{$h$-bad $z \in P$}} \err_P(h) - 1 + |I_\bad(z)|
        \Big) \nn \\
        &\le&
        \Big(\fr{1}{m} \sum_{\text{$h$-good $z \in P$}} err_P(h) - |I_\good(z)| \Big) + \Big(\fr{1}{m} \sum_{\text{$h$-bad $z \in P$}} \err_P(h) + |I_\bad(z)|
        \Big) \nn \\
        &=&
        \err_P(h) + \fr{1}{m} \Big(\sum_{\text{$h$-bad $z \in P$}} |I_\bad(z)| -  \sum_{\text{$h$-good $z \in P$}} |I_\good(z)|\Big)
        \nn \\
        \explain{by \eqref{eqn:rpe:good-bad-equal}}
        &=&
        \err_P(h). \nn
    }
    This completes the inductive step of the proof of Lemma~\ref{lmm:rpe:h-good}.

\extraspacing {\bf Remark.} The approximation ratio 2 is the best possible for \ttt{RPE}. To see this, consider the input $P$ in Figure~\ref{fig:rpe:2tight}, where $n-1$ white points have label $-1$ and the only black point has label 1. The optimal monotone error $k^*$ is 1 (achieved by the monotone classifier that maps all points to $-1$). If \ttt{RPE} probes the black point first --- which happens with probability $1/n$ --- then $h_\rpe$ misclassifies all the white points and, thus, incurs an error of $n - 1$. On the other hand, if the first point probed is white, then $h_\rpe$ misclassies only the black point and incurs an error of $1$. The expected error of $h_\rpe$ is therefore $\fr{1}{n} \cdot (n-1) + (1 - \fr{1}{n}) \cdot 1 = 2 - \fr{2}{n}$, which gets arbitrarily close to 2 as $n$ increases.

\begin{figure}
    \centering
    \hspace{35mm}
    \includegraphics[height=30mm]{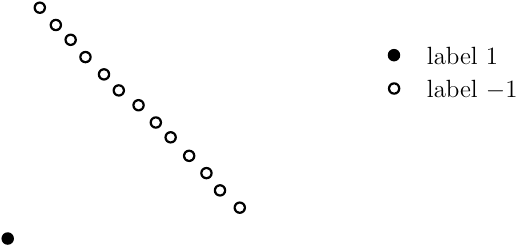}
    \figcapup
    \caption{The approximation ratio 2 is tight for \ttt{RPE}}
    \label{fig:rpe:2tight}
    \figcapdown
\end{figure}

\subsection{The Expected Cost of \ttt{RPE}} \label{sec:rpe:cost}

\extraspacing {\bf Chains.} Let us first review a property of the width $w$ defined in \eqref{eqn:intro:w}. Call a subset $S \subseteq P$ a {{\em chain}} if we can arrange the elements of $S$ into a sequence $p_1 \dombyeq p_2 \dombyeq  ... \dombyeq p_{|S|}$. A {{\em chain}} {{\em decomposition}} of $P$ is a collection of disjoint chains $C_1$, $C_2$, ..., $C_t$ ($t \ge 1$) whose union is $P$. Dilworth's Theorem \cite{d50} states that there must exist a chain decomposition of $P$ containing $w$ chains.

\begin{example}
    The input $P$ in Figure~\ref{fig:intro-ds} can be decomposed into 6 chains: $C_1 = \{p_1, p_2, p_3, p_4, p_{10}\}$, $C_2$ $=$ $\{p_{11}\}$, $C_3 = \{p_5, p_9, p_{12}\}$, $C_4 = \{p_{16}\}$, $C_5 = \{p_{13}\}$, and $C_6 = \{p_6,$ $p_7,$ $p_8, p_{14}, p_{15}\}$. The points $p_{10},$ $p_{11},$ $p_{12},$ $p_{16},$ $p_{13}$, and $p_{14}$ indicate the absence of any chain decomposition of $P$ having less than 6 chains.
\end{example}

\extraspacing {\bf Attrition and Elimination.} Before discussing the cost of \ttt{RPE}, let us take a detour to discuss an {{\em attrition-and-elimination (A\&E) game}} between Alice and Bob. The input is a chain $C$ of $m \ge 1$ points in $\real^d$. In each round:
\begin{itemize}
    \item Bob performs ``attrition'' by either doing nothing or arbitrarily deleting some points from $C$;

    \item Alice then carries out ``elimination'' by picking a point $p \in C$ uniformly at random, and deleting from $C$ all the points $q \domeq p$.
\end{itemize}
The game ends when $C$ becomes empty. The number of rounds is a random variable depending on Bob's strategy. The question is how Bob should play to maximize the expectation of this variable. \rev{The game is a generalization of ``random binary search'', in which Bob always chooses to do nothing in each round.}

\begin{lemma} \label{lmm:rpe:a-and-e}
    Regardless of Bob's strategy, the game has $O(\Log m)$ rounds in expectation.
\end{lemma}

\begin{proof}
    Let $X$ be the number of elements left in $C$ after the first round. We show $\Pr[X \le m/2] > 1/2$. Let $C'$ be the content of $C$ after Bob's attrition in the first round. Set $m' = |C'|$. Let us arrange the points of $C'$ such that $p_1 \dombyeq p_2 \dombyeq  ... \dombyeq p_{m'}$. If Alice picks $p = p_i$ ($i \in [m']$), then at most $i - 1$ points are left after her elimination. Hence, $X \le m'/2$ as long as $i \le 1 + m'/2$, which occurs with probability greater than $1/2$. The fact $\Pr[X \le m/2] > 1/2$ now follows from $m' \le m$.

    \vgap

    We call a round {\em successful} if it reduces the number of elements in $C$ by at least a factor of 2. The total number of successful rounds cannot be more than $\log_2 (1+m)$. Let $Y$ be the total number of rounds. If $Y \ge 2 \log_2 (1+m)$, there must be at least $\log_2 (1+m)$ unsuccessful rounds; as all the rounds are independent, this can happen with probability less than $(1/2)^{\log_2 (1+m)} < 1/m$. Because $Y$ is trivially bounded by $m$, we have $\expt[Y] \le 2 \log_2 (1+m) + m \cdot \Pr[\text{at least $2 \log_2 (1+m)$ rounds}] = O(\Log m)$.
\end{proof}

\noindent {\bf Cost Analysis of RPE.} Returning to \ttt{RPE}, let $\{C_1, C_2, ..., C_w\}$ be an arbitrary chain decomposition of $P$ with $w$ chains. Note that \ttt{RPE} is unaware of these chains. For each $i \in [w]$, break the chain $C_i$ into two disjoint subsets:
\myitems{
    \item $C^\pos_i = \set{p \in C_i \mid \lab(p) = 1}$, and
    \item $C^\neg_i = \set{p \in C_i \mid \lab(p) = -1}$.
}

\begin{lemma} \label{lmm:rpe:cost-one-chain}
    \ttt{RPE} probes $O(\Log |C^\pos_i|)$ points from $C^\pos_i$ in expectation.
\end{lemma}

\begin{proof}
    The operations that \ttt{RPE} performs on $C^\pos_i$ can be modeled as an A\&E game on an initial input $C = C^\pos_i$, as explained next.

    \vgap

    In each round (of the A\&E game), Bob formulates his strategy according to the execution of \ttt{RPE}. Suppose that \ttt{RPE} probes an element outside $C^\pos_i$ at Line 3 (of the pseudocode in Section~\ref{sec:intro:techniques}). Recall that the algorithm removes some elements from $P$ at Line 4 or 5. Accordingly, Bob carries out attrition by deleting from $C$ all the elements of $C^\pos_i$ removed by \ttt{RPE}. After that, Bob observes the next probe of \ttt{RPE} and performs attrition in the same way as long as the element probed is outside $C_i$. If, on the other hand, \ttt{RPE} probes an element $z \in C^\pos_i$, he passes the turn to Alice.

    \vgap

    From Alice's perspective, conditioned on $z \in C^\pos_i$, \ttt{RPE} must have chosen $z$ uniformly at random from the current $C$, namely, the set of elements from $C^\pos_i$ still in $P$. Hence, Alice can take $z$ as her choice in the A\&E game to perform elimination. Because $z$ has label 1, after probing $z$ at Line 3, \ttt{RPE} shrinks $P$ by removing at Line 4 every element $p \domeq z$. As far as $C$ is concerned, the shrinking deletes elements from $C$ in exactly the way Alice should do in her elimination. This completes a round of the A\&E game. The next round starts and proceeds in the same fashion.

    \vgap

    By Lemma~\ref{lmm:rpe:a-and-e}, the A\&E game lasts for $O(\Log |C^\pos_i|)$ rounds in expectation regardless of Bob's strategy. Hence, \ttt{RPE} probes $O(\Log |C^\pos_i|)$ elements from $C^\pos_i$ in expectation.
\end{proof}

By a symmetric argument, \ttt{RPE} probes $O(\Log |C^\neg_i|)$ elements from $C^\neg_i$ in expectation. Therefore, the expected number of elements probed by \ttt{RPE} in total is given by
\myeqn{
    O\left(\sum_{i=1}^w \Log|C^\pos_i| + \Log |C^\neg_i| \right)  =
    O\left(\sum_{i=1}^w \Log|C_i| \right)
    =
    O\left(w  \Log \fr{n}{w} \right)
    \nn
}
where the last step used $\sum_{i=1}^w |C_i| = \rev{n}$ \rev{and the fact that $\sum_{i=1}^w \Log|C_i|$ is \revvv{maximized} when all chains have the same size (\revvv{a standard application of Jensen's inequality}\footnote{\revvv{
The discrete version of Jensen's inequality states that, for a concave function $f$ and any values $x_1,...,x_w$ in its domain, $\frac{1}{w}\sum_{i=1}^{w}f(x_i) \leq f\left(\frac{1}{w}\sum_{i=1}^{w}x_i\right)$. Applying this inequality to the concave function $f(x)=\Log x=\log_2(1+x)$ and $x_i=|C_i|>0$ yields $\frac{1}{w}\sum_{i=1}^{w}\Log|C_i| \leq \Log\left(\frac{1}{w}\sum_{i=1}^{w}|C_i|\right) =\Log(n/w)$. Equality holds when $|C_1|=...=|C_w|=n/w$.
}}).
This completes the proof of Theorem~\ref{thm:rpe}.

\subsection{Application to Monotonicity Testing} \label{sec:rpe:testing}

We finish this section with a remark on monotonicity testing. As reviewed in Section~\ref{sec:intro:prev}, given a multiset $P$ of $n$ points in $\real^d$ and a parameter $\xi \in (0, 1)$, the output of monotonicity testing is
\myitems{
    \item always ``yes'' if $P$ is monotone;
    \item ``no'' with probability at least $2/3$ if $k^* \ge \xi n$ where $k^*$ is the optimal monotone error of $P$ (see \eqref{eqn:k-star});
    \item either ``yes'' or ``no'' if $0 < k^* < \xi n$.
}
Consider the following simple algorithm:
\mytab{
    \>\> 1.\> run \ttt{RPE} on $P$ and obtain $h_\rpe$ from \eqref{eqn:rpe-classifier} \\
    \>\> 2.\> take a set $S$ of $2/\xi$ uniform samples of $P$ with replacement and obtain their labels \\
    \>\> 3.\> {\bf if} $h_\rpe$ misclassifies any element in $S$ {\bf then return} ``no'' \\
    \>\> 4.\> {\bf else return}  ``yes''
}
By Theorem~\ref{thm:rpe}, the algorithm probes $O(w \Log \fr{n}{w}) + 2/\xi$ elements of $P$ in expectation. Next, we explain why it fulfills the output requirements of monotonicity testing. First, if $P$ is monotone, then the algorithm definitely outputs ``yes'' because, as mentioned before, $h_\rpe$ is guaranteed to classify all elements of $P$ correctly in this case. On other hand hand, assume that $k^* \ge \xi n$. Because $\err_P(h_\rpe) \ge k^*$, the probability for $h_\rpe$ to misclassify a uniformly random element of $P$ is at least $k^* / n \ge \xi$. Hence, the probability for $h_\rpe$ to be correct on all the elements in $S$ is at most $(1 - \xi)^{2/\xi} < 1/e^2 < 1/3$. This means that the algorithm outputs ``no'' with probability at least $2/3$.

\section{Relative-Comparison Coresets} \label{sec:coreset}

This section solves Problem 1 up to an approximation ratio of $1 + \eps$ w.h.p.\ assuming $\eps \le 1$ (for $\eps > 1$, reset it to 1). The central step is to find a relative-comparison coreset of the input $P$. Recall from Section~\ref{sec:intro:techniques} that this is a subset $Z \subseteq P$ where every element has its label revealed and carries a positive weight such that the weighted error of every monotone classifier $h$ on $Z$ --- namely, $\werr_Z(h)$ defined in \eqref{eqn:intro:weighted-err} --- satisfies the condition in \eqref{eqn:intro:F-property-thru-coreset} (remember that the $\Delta$ value in \eqref{eqn:intro:F-property-thru-coreset} may be unknown). Formally, we establish:

\begin{theorem} \label{thm:coreset}
    Let $n$ and $w$ be the size and width of the input $P$ to Problem 1, respectively. In $O(\fr{w}{\eps^2} \Log \fr{n}{w} \cdot \log \fr{n}{\delta})$ probes, we can obtain with probability at least $1-\delta$ a relative-comparison coreset $Z$ of $P$ with size $|Z| = O(\fr{w}{\eps^2} \Log \fr{w}{n} \cdot \log \fr{n}{\delta})$.
\end{theorem}

As explained in Section~\ref{sec:intro:techniques}, given the coreset $Z$ in Theorem~\ref{thm:coreset}, we can solve Problem 1 by finding a monotone classifier $h^\newstar$ minimizing $\werr_Z(h^\newstar)$. This requires no more probing and can be done in CPU time polynomial in $|Z|$ and $d$ (see \cite{ahkw06,s13b}). This leads us to:

\begin{corollary} \label{crl:coreset:prob1}
    For Problem 1, there is an algorithm that finds w.h.p.\ a monotone classifier with an error at most $(1+\eps) k^*$ by probing $O(\fr{w}{\eps^2} \Log \fr{w}{n} \cdot \log n)$ elements, where $n$ is the size of the input $P$, $w$ is its width, and $k^*$ is its optimal monotone error.
\end{corollary}

The rest of the section serves as a proof of Theorem~\ref{thm:coreset}. The main difficulty arises from establishing its correctness for $d = 1$. Most of our discussion revolves around the following problem.

\boxminipg{0.95\linewidth}{
\begin{problem}
    Let $P$ be a multiset of 1D labeled points as defined in Problem 1 ($d = 1$), and $\eps$ be a value in $(0, 1]$. Find a function $F: \mono \rightarrow \real$ such that every $h \in \mono$ satisfies the following two inequalities:
     \myeqn{
        |F(h) - \err_P(h)| \le \eps |P| / 64 \label{eqn:coreset:F-property1} \\
        \err_P(h) \cdot \left(1-\fr{\eps}{4}\right) + \Delta
        \le
        F(h)
        \le
        \err_P(h) \cdot \left(1+\fr{\eps}{4}\right) + \Delta
        \label{eqn:coreset:F-property2}
    }
    where $\Delta$ is a possibly unknown value with
    \myeqn{
        |\Delta| &\le&  \eps |P| / 64. \label{eqn:coreset:Delta-property}
    }
    The efficiency of an algorithm is measured by the number of elements probed.
\end{problem}
}

Sections~\ref{sec:coreset:1d-warmup}-\ref{sec:coreset:1d-alg} settles Problem 2 with probability at least $1-\delta$ by probing $O(\fr{1}{\eps^2} \Log \fr{1}{n} \cdot \log \fr{n}{\delta})$ elements. Section~\ref{sec:coreset:1d-coreset} then utilizes our solution to build a relative-comparison coreset that meets the requirements of Theorem~\ref{thm:coreset} for $d = 1$. Section~\ref{sec:coreset:any-d}  extends the discussion to $d > 1$.

\subsection{Warm Up: A Special Case} \label{sec:coreset:1d-warmup}

Solving Problem 2 requires proving the existence of $\Delta$ \rev{without deciding its value explicity.} This subsection illustrates the principle in the special case where all elements of $P$ have an identical value. Any monotone classifier $h$ maps the entire $P$ to 1 or $-1$. Denote by $h^\pos$ (resp., $h^\neg$) the monotone classifier that always outputs 1 (resp., $-1$). It suffices to construct a function $F: \set{h^\pos, h^\neg} \rightarrow \real$ that satisfies \eqref{eqn:coreset:F-property1}-\eqref{eqn:coreset:Delta-property} for $h \in \set{h^\pos, h^\neg}$.

\vgap

\rev{It suffices to} return an arbitrary function $F: \set{h^\pos, h^\neg} \rightarrow \real$  satisfying \eqref{eqn:coreset:F-property1}. It is standard to construct such a function via random sampling; the details are given in Section~\ref{sec:coreset:1d-alg}. \rev{Next, we argue for the existence of $\Delta$ that meets} the requirements in \eqref{eqn:coreset:F-property2} and \eqref{eqn:coreset:Delta-property}. W.l.o.g., assume that the optimal monotone error $k^*$ is achieved by $h^\pos$, namely, $k^* = \err_P(h^\pos)$. Our argument distinguishes two scenarios depending on whether $k^*$ is large.

\extraspacing {\bf When $\bm{k^* \ge |P| / 16}$.} In this scenario, we set $\Delta = 0$, which trivially satisfies \eqref{eqn:coreset:Delta-property}. To prove \eqref{eqn:coreset:F-property2}, recall that $F(h)$ estimates $\err_P(h)$ up to an \rev{additive} offset of $\eps |P|/64$ for $h \in \set{h^\pos, h^\neg}$. Since $\err_P(h) \ge |P|/16$, the offset is at most $\fr{\eps}{4} \err_P(h)$. This yields  $\err_P(h) \cdot (1-\fr{\eps}{4})         \le F(h) \le \err_P(h) \cdot (1+\fr{\eps}{4})$.

\extraspacing {\bf When $\bm{k^* < |P| / 16}$.} In this scenario, we set
\myeqn{
    \Delta &=& F(h^\pos) - k^*. \label{eqn:coreset:1d-warmup-Delta}
}
\rev{Note that $\Delta$ is not computable because $k^*$ is unknown.} Furthermore, as $F$ satisfies \eqref{eqn:coreset:F-property1} for $h = h^\pos$, we have $|\Delta| = \rev{|F(h^\pos) - k^*| = |F(h^\pos) - \err_P(h^\pos)|}  \le \eps|P| / 64$; hence, \eqref{eqn:coreset:Delta-property} holds.

\vgap

Next, we prove \eqref{eqn:coreset:F-property2} for $h \in \set{h^\pos, h^\neg}$. The case $h = h^\pos$ is immediate because \eqref{eqn:coreset:F-property2} reduces to $k^* (1-\eps/4) \le k^* \le k^* (1+\eps/4)$, which is clearly true. For $h^\neg$, since $k^* < |P|/16$, we have $\err_P(h^\neg) = |P| - \err_P(h^\pos) \ge \fr{15}{16} |P|$. Thus, from \eqref{eqn:coreset:F-property1}:
\myeqn{
    |F(h^\neg) - \err_P(h^\neg)| \le \fr{\eps |P|}{64} \le \fr{\eps}{60} \cdot \err_P(h^\neg). \nn
}
On the other hand, as explained earlier, \rev{$|\Delta| \le \fr{\eps |P|}{64}$, which yields $|\Delta| \le \fr{\eps}{60} \cdot \err_P(h^\neg)$.}
Hence
\myeqn{
    |F(h^\neg) - \err_P(h^\neg) - \Delta|
    \le
     \fr{\eps}{60} \cdot \err_P(h^\neg) + \fr{\eps}{60} \cdot \err_P(h^\neg)
    <
    \fr{\eps}{4} \cdot \err_P(h^\neg). \nn
}
We thus conclude that \eqref{eqn:coreset:F-property2} holds for $h = h^\neg$.

\subsection{A Recursive Framework for Problem 2} \label{sec:coreset:1d-framework}

This subsection discusses Problem 2 in its generic setting. \rev{If $P = \emptyset$, we return $F(h) = 0$ for all $h \in \mono$; if $|P| = 1$, we return
\myeqn{
    F(h) = \err_P(h) \label{eqn:coreset:F-for-n-equas-1}
}
for all $h \in \mono$. In both cases, the function $F$ satisfies \eqref{eqn:coreset:F-property1}-\eqref{eqn:coreset:Delta-property} with $\Delta = 0$}.

\vgap

Our subsequent discussion assumes $|P| \ge 2$. When $d = 1$, a monotone classifier $h$ takes the form
\myeqn{
    h(p) &=& \left\{
    \begin{tabular}{ll}
        1 & if $p > \tau$ \\
        $-1$ & otherwise
    \end{tabular}
    \right. \label{eqn:coreset:1d-classifier}
}
which is parameterized by a value $\tau$. We sometimes make the parameter explicit by representing the classifier in \eqref{eqn:coreset:1d-classifier} as $h^\tau$.

\vgap

Our framework assumes the availability of a function $G_1: \mono \rightarrow \real$ \rev{having two properties:}
\rev{
\myitems{
    \item {{\bf G1-1}:} $G_1$ approximates $\err_P$ up to an absolute error of $\eps |P| / 64$, namely,
    \myeqn{
        |G_1(h) - \err_P(h)|
        &\le&
        \eps |P| / 64
        \label{eqn:coreset:g1-gaurantee}
    }
for all $h \in \mono$.
    \item {{\bf G1-2}:} for any $\tau \in \real$, it holds that $G_1(h^\tau) = G_1(h^{\tau'})$ where $\tau'$ is the predecessor of $\tau$ in $P$ (i.e., $\tau'$ is the largest element of $P$ that is at most $\tau$); specially, if $\tau'$ does not exist, then $G_1(h^\tau) = G_1(h^{-\infty})$ (note: $h^{-\infty}$ is simply $h^\pos$).
}
}

Define
\myeqn{
    \alpha &=& \text{the smallest $\rev{\tau \in P \cup \set{-\infty}}$ with $G_1(h^\tau) < |P| \cdot \left(\fr{1}{4} - \fr{\eps}{64} \right)$} \label{eqn:coreset:alpha} \\
    \beta &=& \text{the largest $\rev{\tau \in P \cup \set{-\infty}}$ with $G_1(h^\tau) < |P| \cdot \left(\fr{1}{4} - \fr{\eps}{64} \right)$}. \label{eqn:coreset:beta}
}
\rev{If neither $\alpha$ nor $\beta$ exists --- which occurs when $G_1(h) \ge |P| (\fr{1}{4} - \fr{\eps}{64})$ for all $h \in \mono$ due to property {\bf G1-2} --- define $\alpha = \beta = \ttt{null}$. Note that either $\alpha$ and $\beta$ both exist, or neither does.}

\vgap

Break $P$ into:
\myeqn{
    P_\alpha &=& \left\{
        \begin{tabular}{ll}
            $\emptyset$ & if $\alpha = \beta = \ttt{null}$ \\
            $\{ \text{$p \in P$} \mid p = \alpha \}$ & otherwise
        \end{tabular}
    \right.
    \label{eqn:coreset:P_alpha} \\
    P_\middle &=&
    \left\{
        \begin{tabular}{ll}
            $\emptyset$ & if $\alpha = \beta = \ttt{null}$ \\
            $\{ \text{$p \in P$} \mid \alpha < p \le \beta\}$ & otherwise
        \end{tabular}
    \right.
    \label{eqn:coreset:P_middle} \\
    P_\rest &=& \rev{P \setm (P_\alpha \cup P_\middle)}. \label{eqn:coreset:P_rest}
}
These are multisets where each point inherits its label in $P$.

\begin{proposition} \label{prop:coreset:P_middle-size}
    $|P_\middle| < |P|/2$.
\end{proposition}
\begin{proof}
    \rev{This is trivial if $\alpha = \beta = \ttt{null}$. Next, we consider that both $\alpha$ and $\beta$ exist.}

    \vgap

    $P_\middle$ must have less than $|P|/4$ elements of label 1. To see why, assume that $P$ has at least $|P|/4$ elements in $(\alpha, \beta]$ having label 1. Thus, $\err_P(h^\beta) \ge |P|/4$, which together with \eqref{eqn:coreset:g1-gaurantee} tells us $G_1(h^\beta) \ge |P| (\fr{1}{4} - \fr{\eps}{64})$, contradicting the definition of $\beta$.
    Similarly, $P_\middle$ must have less than $|P|/4$ elements of label $-1$. To see why, assume that $P$ has at least $|P|/4$ elements in $(\alpha, \beta]$ having label $-1$. Thus, $\err_P(h^\alpha) \ge |P|/4$, which together with \eqref{eqn:coreset:g1-gaurantee} tells us $G_1(h^\alpha) \ge |P| (\fr{1}{4} - \fr{\eps}{64})$, contradicting the definition of $\alpha$.
\end{proof}

We also assume the availability of another function $ G_2: \mono \rightarrow \real$ having two properties:
\myitems{
    \item {{\bf G2-1}:} $G_2$ approximates $\err_{P_\rest}$ up to absolute error $\eps |P_\rest| / 64$, namely, for any $h \in \mono$
    \myeqn{
        |G_2(h) - \err_{P_\rest}(h)|
        &\le&
        \eps |P_\rest| / 64;
        \label{eqn:coreset:g2-gaurantee1}
    }

    \item {{\bf G2-2}} \rev{(applicable only if $\alpha$ and $\beta$ exist):} \rev{Define
    \myeqn{
        \beta_\mit{next}
        =
        \text{the smallest element of $P$ greater than $\beta$ or $\infty$ if no such element exists.} \label{eqn:coreset:beta-next}
    }
    For any $\tau \in [\alpha, \beta_\mit{next})$,} it holds that
    \myeqn{
        G_2(h^\tau) = G_2(h^\beta). \label{eqn:coreset:g2-guarantee2}
    }
}
By solving Problem 2 on $P_\alpha$ using the solution in Section~\ref{sec:coreset:1d-warmup} \rev{(all the elements in $P_\alpha$ have an identical value, which is $\alpha$)} and on $P_\middle$ recursively, we obtain functions $F_\alpha: \mono \rightarrow \real$ and $F_\middle: \mono \rightarrow \real$ such that every $h \in \mono$ satisfies \eqref{eqn:coreset:f_alpha-property1}-\eqref{eqn:coreset:f_mid-property2}:
\myeqn{
    |F_\alpha(h) - \err_{P_\alpha}(h)| \le \eps |P_\alpha| / 64
    \label{eqn:coreset:f_alpha-property1} \\
    |F_\middle(h) - \err_{P_\middle}(h)| \le \eps |P_\middle| / 64
    \label{eqn:coreset:f_mid-property1} \\
    \err_{P_\alpha}(h) \cdot \left(1-\fr{\eps}{4}\right) + \Delta_\alpha
    \le
    F_\alpha(h)
    \le
    \err_{P_\alpha}(h) \cdot \left(1+\fr{\eps}{4}\right) + \Delta_\alpha
    \label{eqn:coreset:f_alpha-property2} \\
    \err_{P_\middle}(h) \cdot \left(1-\fr{\eps}{4}\right) + \Delta_\middle
    \le
    F_\middle(h)
    \le
    \err_{P_\middle}(h) \cdot \left(1+\fr{\eps}{4}\right) + \Delta_\middle
    \label{eqn:coreset:f_mid-property2}
}
where $\Delta_\alpha$ and $\Delta_\middle$ are (unknown) real values such that
\myeqn{
    |\Delta_\alpha| \le \eps |P_\alpha| / 64
    \label{eqn:coreset:f_alpha-Delta-property} \\
    |\Delta_\middle| \le \eps |P_\middle| / 64
    \label{eqn:coreset:f_mid-Delta-property}
}
The target function $F$ for Problem 2 can now be finalized as
\myeqn{
    F(h) =
    G_2(h) + F_\alpha(h) + F_\middle(h).
    \label{eqn:coreset:F-case2}
}
\rev{As a remark, if $\alpha = \beta = \ttt{null}$, then $F_\alpha(h) = F_\middle(h) = 0$, leaving $F(h) = G_2(h)$.}

\subsection{Correctness of the Framework} \label{sec:coreset:1d-correctness}

Next, we prove that the above framework always produces a function $F$ obeying \eqref{eqn:coreset:F-property1}-\eqref{eqn:coreset:Delta-property}. \rev{The base case where $|P| = 0$ or $1$ has been resolved in Section~\ref{sec:coreset:1d-framework}}. Inductively, assuming that \eqref{eqn:coreset:f_mid-property1}, \eqref{eqn:coreset:f_mid-property2}, and \eqref{eqn:coreset:f_mid-Delta-property} hold on $P_\middle$, we show that the function $F$ in \eqref{eqn:coreset:F-case2} satisfies \eqref{eqn:coreset:F-property1}-\eqref{eqn:coreset:Delta-property} with
\myeqn{
    \Delta
    &=&
    \left\{
    \begin{tabular}{ll}
        0 & if $\alpha = \beta = \ttt{null}$ \\
        $\Delta_\alpha + \Delta_\middle + G_2(h^{\beta}) - \err_{P_\rest}(h^{\beta})$ & otherwise
    \end{tabular}
    \right.
    \label{eqn:correctness-Delta-case2}
}

\extraspacing {\bf Proof of \eqref{eqn:coreset:F-property1}.} For any $h \in \mono$, it holds that
\myeqn{
    \err_P(h)
    &=&
    \err_{P_\alpha}(h)
    +
    \err_{P_\middle}(h)
    +
    \err_{P_\rest}(h) .
    \label{eqn:err-case2}
}
Combining the above with \eqref{eqn:coreset:F-case2} gives:
\myeqn{
    |F(h) - \err_P(h)|
    &\le&
    |F_\alpha(h) - \err_{P_\alpha}(h)|
    +
    |F_\middle(h) - \err_{P_\middle}(h)|
    +
    |G_2(h) - \err_{P_\rest}(h)| \nn \\
    \textrm{(by \eqref{eqn:coreset:g2-gaurantee1}, \eqref{eqn:coreset:f_alpha-property1}, and \eqref{eqn:coreset:f_mid-property1})}
    &\le&
    \fr{\eps |P_\alpha|}{64} + \fr{\eps |P_\middle|}{64} + \fr{\eps |P_\rest|}{64} \nn \\
    &\le&
    \fr{\eps |P|}{64}.
    \nn
}

\extraspacing {\bf Proof of \eqref{eqn:coreset:Delta-property}.} \rev{If $\alpha = \beta = \ttt{null}$, then $\Delta = 0$ trivially satisfies \eqref{eqn:coreset:Delta-property}. Otherwise,} from \eqref{eqn:correctness-Delta-case2}, \eqref{eqn:coreset:g2-gaurantee1}, \eqref{eqn:coreset:f_alpha-Delta-property}, and \eqref{eqn:coreset:f_mid-Delta-property}, we know
\myeqn{
    |\Delta| = |\Delta_\alpha + \Delta_\middle + G_2(h^{\beta}) - \err_{P_\rest}(h^{\beta})|
    \le
    \fr{\eps |P_\alpha|}{64} + \fr{\eps |P_\middle|}{64} + \fr{\eps |P_\rest|}{64} 
    \le
    \fr{\eps |P|}{64} \label{eqn:coreset:correctness-bound-on-Delta}. 
}

\extraspacing {\bf Proof of \eqref{eqn:coreset:F-property2}.} \rev{We first present a useful lemma.}

\begin{lemma} \label{lmm:coreset:F-property2:h1}
    \rev{$F(h)$ satisfies \eqref{eqn:coreset:F-property2} whenever $G_1(h) \ge |P|(\fr{1}{4} - \fr{\eps}{64})$. }
\end{lemma}

\begin{proof}
    \rev{From \eqref{eqn:coreset:g1-gaurantee} and $G_1(h) \ge |P|(\fr{1}{4} - \fr{\eps}{64})$, we have
    \myeqn{
        \err_P(h)
        \ge
        G_1(h) - \fr{\eps |P|}{64}
        \ge
        \fr{|P|}{4} - \fr{\eps |P|}{64} - \fr{\eps |P|}{64}
        \ge
        \fr{14|P|}{64}
        \label{eqn:coreset:correctness-1}
    }
    where the last step used $\eps \le 1$.
    Combining the above with \eqref{eqn:coreset:correctness-bound-on-Delta} yields
    \myeqn{
        \fr{\eps \cdot \err_P(h)}{4} + \Delta
        \ge
        \fr{\eps}{4} \cdot \fr{14 |P|}{64} - \fr{\eps |P|}{64}
        =
        \fr{10 \eps |P|}{256}. \nn
    }
    As proved earlier, $F$ satisfies \eqref{eqn:coreset:F-property1}; hence:
    \myeqn{
        F(h) - \err_P(h)
        \le
        \fr{\eps |P|}{64}
        <
        \fr{10 \eps |P|}{256}
        \le
        \fr{\eps \cdot \err_P(h)}{4} + \Delta. \label{eqn:coreset:correctness:help1}
    }
    Similarly, from \eqref{eqn:coreset:correctness-bound-on-Delta} and \eqref{eqn:coreset:correctness-1}, we know
    \myeqn{
        \fr{\eps \cdot \err_P(h)}{4} - \Delta
        \ge
        \fr{\eps}{4} \cdot \fr{14 |P|}{64} - \fr{\eps |P|}{64}
        =
        \fr{10 \eps |P|}{256}. \nn
    }
    Hence, \eqref{eqn:coreset:F-property1} tells us
    \myeqn{
        \err_P(h) - F(h)
        \le
        \fr{\eps |P|}{64}
        <
        \fr{10 \eps |P|}{256}
        \le
        \fr{\eps \cdot \err_P(h)}{4} - \Delta. \label{eqn:coreset:correctness:help2}
    }
    The correctness of \eqref{eqn:coreset:F-property2} now follows from \eqref{eqn:coreset:correctness:help1} and \eqref{eqn:coreset:correctness:help2}.}
\end{proof}

\rev{As a corollary, when $\alpha = \beta = \ttt{null}$ (namely, $G_1(h) \ge |P| (\fr{1}{4} - \fr{\eps}{64})$ for all $h \in \mono$), $F(h)$ satisfies \eqref{eqn:coreset:F-property2} for all $h \in \mono$.}

\vgap

\rev{Now, we consider that $\alpha$ and $\beta$ exist. By property {\bf G1-2} and the definitions of $\alpha$ and $\beta$, we know $G_1(h^\tau) \ge |P| (\fr{1}{4} - \fr{\eps}{64})$ for any $\tau$ satisfying $\tau < \alpha$ and $\tau \ge \beta_\mit{next}$ (the reader may want to revisit the definitions of $\alpha, \beta$, and $\beta_\mit{next}$ in \eqref{eqn:coreset:alpha}, \eqref{eqn:coreset:beta}, and \eqref{eqn:coreset:beta-next}). By Lemma~\ref{lmm:coreset:F-property2:h1}, $F(h^\tau)$ must satisfy \eqref{eqn:coreset:F-property2} for those $\tau$ values. It remains to prove that this is also true for $\tau \in [\alpha, \beta_\mit{next})$, as we show next.}

\begin{lemma}
    \rev{If $\alpha$ and $\beta$ exist, $F(h^\tau)$ satisfies
    \eqref{eqn:coreset:F-property2} for $\tau \in [\alpha, \beta_\mit{next})$.}
\end{lemma}

\begin{proof}
    For any $\tau \in [\alpha, \rev{\beta_\mit{next})}$, we have
    \myeqn{
        \err_{P_\rest}(h^\tau)
        &=& \err_{P_\rest}(h^\beta)
        \label{eqn:correctness-2}
    }
    because $P_\rest$ has no element in $[\alpha, \rev{\beta_\mit{next})}$. By property {\bf G2-2}, $G_2(h^\tau) = G_2(h^\beta)$ for all $\tau \in [\alpha, \rev{\beta_\mit{next})}$. This, together with \eqref{eqn:coreset:F-case2}, yields
    \myeqn{
        F(h^\tau) = G_2(h^{\beta}) + F_\alpha(h^\tau) + F_\middle(h^\tau).
        \label{eqn:coreset:F-rewritten-with-G2}
    }

    \vgap

    We can thus derive
    \myeqn{
        && \err_P(h^\tau) (1+\eps/4) + \Delta  \nn \\
        \textrm{(by \eqref{eqn:err-case2} and \eqref{eqn:correctness-2})}
        &=&
        (\err_{P_\alpha}(h^\tau) + \err_{P_\middle}(h^\tau) + \err_{P_\rest}(h^\beta)) (1+\eps/4) + \Delta \nn \\
        &\ge&
        (\err_{P_\alpha}(h^\tau) + \err_{P_\middle}(h^\tau)) (1+ \eps/4) + \err_{P_\rest}(h^\beta) + \Delta \nn \\
        \textrm{(by \eqref{eqn:correctness-Delta-case2})}
        &=&
        (\err_{P_\alpha}(h^\tau) + \err_{P_\middle}(h^\tau)) (1+ \eps/4) + \Delta_\alpha + \Delta_\middle + G_2(h^{\beta}) \nn \\
        \textrm{(by \eqref{eqn:coreset:f_alpha-property2} and \eqref{eqn:coreset:f_mid-property2})}
        &\ge&
        F_\alpha(h^\tau) +
        F_\middle(h^\tau) + G_2(h^{\beta}) \nn \\
        \explain{by \eqref{eqn:coreset:F-rewritten-with-G2}}
        &=& F(h^\tau). \nn
    }
    Similarly,
    \myeqn{
        && \err_P(h^\tau) (1-\eps/4) + \Delta  \nn \\
        \textrm{(by \eqref{eqn:err-case2} and \eqref{eqn:correctness-2})}
        &=&
        (\err_{P_\alpha}(h^\tau) + \err_{P_\middle}(h^\tau) + \err_{P_\rest}(h^\beta)) (1 - \eps/4) + \Delta \nn \\
        &\le&
        (\err_{P_\alpha}(h^\tau) + \err_{P_\middle}(h^\tau)) (1- \eps/4) + \err_{P_\rest}(h^\beta)  + \Delta \nn \\
        \textrm{(by \eqref{eqn:correctness-Delta-case2})}
        &=&
        (\err_{P_\alpha}(h^\tau) + \err_{P_\middle}(h^\tau)) (1- \eps/4) + \Delta_\alpha + \Delta_\middle + G_2(h^{\beta}) \nn \\
        \textrm{(by \eqref{eqn:coreset:f_alpha-property2} and \eqref{eqn:coreset:f_mid-property2})}
        &\le&
        F_\alpha(h^\tau) + F_\middle(h^\tau) + G_2(h^{\beta}) \nn  \\
        &=&
        F(h^\tau). \nn
    }
    This completes the proof.
\end{proof}

\subsection{A Concrete Algorithm for Problem 2} \label{sec:coreset:1d-alg}

Instantiating our framework in Section~\ref{sec:coreset:1d-framework}  requires constructing the function $F$ in Section~\ref{sec:coreset:1d-warmup} and \rev{the functions $G_1$ and $G_2$ in Section~\ref{sec:coreset:1d-framework}}. We explain how to do so in this subsection, by factoring in the consideration that the framework needs to succeed with probability at least $1 - \delta$. Denote by $\ell$ the number of recursion levels; the value $\ell$ is $O(\log n)$ due to Proposition~\ref{prop:coreset:P_middle-size}.

\extraspacing {\bf Constructing $\bm{G_1}$ and $\bm{G_2}$.} Both $G_1$ and $G_2$ map $\mono$ to $\real$. Although $\mono$ has an infinite size, there exists a finite set of ``effective'' classifiers:
\myeqn{
    \mono^P
    &=&
    \set{h^\tau \mid \tau \in P \textrm{ or } \tau = -\infty}. \nn
}
Every monotone classifier has the same error on $P$ as a classifier in $\mono^P$.

\vgap

To build $G_1$, we uniformly sample with replacement a set $S_1$ of $O(\fr{1}{\eps^2} \log \fr{|P| \ell}{\delta})$ elements from $P$. For each $h \in \mono$, define
\myeqn{
    G_1(h)
    &=&
    \fr{|P|}{|S_1|} \cdot \err_{S_1}(h).
    \label{eqn:coreset:G1-constructed}
}
As discussed in Appendix~\ref{app:abs-est}, $G_1(h)$ satisfies \eqref{eqn:coreset:g1-gaurantee} with probability at least $1 - \fr{\delta}{3 \ell \cdot (|P|+1)}$ for each $h \in \mono^P$. As $|\mono^P| \le |P|+1$, $G_1$ satisfies \eqref{eqn:coreset:g1-gaurantee} for all $h \in \mono^P$ --- thus for all $h \in \mono$ --- with probability at least $1 - \delta/(3 \ell)$. \rev{It is clear that $G_1$ has property {\bf G1-2}.}

\vgap

\rev{If $P_\rest$ --- defined in \eqref{eqn:coreset:P_rest} --- is empty, we set $G_2(h) = 0$ for all $h \in \mono$. Otherwise,} we uniformly sample with replacement a set $S_2$ of $O(\fr{1}{\eps^2} \log \fr{|P| \ell}{\delta})$ elements from $P_\rest$. For each $h \in \mono$, define
\myeqn{
    G_2(h)
    &=&
    \fr{|P_\rest|}{|S_2|} \cdot \err_{S_2}(h).
    \label{eqn:coreset:G2-constructed}
}
An argument analogous to the one used earlier for $G_1$ shows that $G_2$ obeys \eqref{eqn:coreset:g2-gaurantee1} for all $h  \in \mono$ with probability at least $1 - \delta/(3 \ell)$. $G_2$ has property {\bf G2-2} because $S_2$ has no elements in $[\alpha, \rev{\beta_\mit{next})}$, \rev{where $\alpha$ and $\beta_\mit{next}$ are defined in \eqref{eqn:coreset:alpha} and \eqref{eqn:coreset:beta-next}, respectively}.

\extraspacing {\bf Constructing the Function $\bm{F}$ in Section~\ref{sec:coreset:1d-warmup}.} Our framework applies the method in Section~\ref{sec:coreset:1d-warmup} to solve Problem 2 on a non-empty $P_\alpha$ --- defined in \eqref{eqn:coreset:P_alpha} --- whose goal is to obtain a function $F_\alpha$ satisfying \eqref{eqn:coreset:f_alpha-property1}. To build $F_\alpha$ (a.k.a.\ the function $F$ in Section~\ref{sec:coreset:1d-warmup} when $P = P_\alpha$), we uniformly sample with replacement a set $S_\alpha$ of $O(\fr{1}{\eps^2} \log \fr{|P| \ell}{\delta})$ elements from $P_\alpha$. For each $h \in \mono$, define
\myeqn{
    F_\alpha(h)
    &=&
    \fr{|P_\alpha|}{|S_\alpha|} \cdot \err_{S_\alpha}(h).
    \label{eqn:coreset:F_alpha-constructed}
}
It satisfies \eqref{eqn:coreset:f_alpha-property1} for all $h  \in \mono$ with probability at least $1 - \delta/(3 \ell)$.

\extraspacing {\bf Putting All Levels Together.} In summary, at each recursion level, by probing $O(\fr{1}{\eps^2} \log \fr{|P| \ell}{\delta})$ elements we can construct the desired functions $G_1$, $G_2$, and $F_\alpha$ with probability at least $1 - \delta / \ell$. As there are $\ell$ levels, the overall cost is $O(\fr{\ell}{\eps^2} \log \fr{n}{\delta}) = O(\fr{\log n}{\eps^2} \cdot \log \fr{n}{\delta})$, and we solve Problem 2 with probability at least $1-\delta$.

\subsection{A One-Dimensional Relative-Comparison Coreset} \label{sec:coreset:1d-coreset}

We are ready to prove Theorem~\ref{thm:coreset} for $d = 1$. Let us re-examine our algorithm (combining Sections~\ref{sec:coreset:1d-warmup}, \ref{sec:coreset:1d-framework}, and \ref{sec:coreset:1d-alg}) and construct a coreset $Z$ along the way.
\rev{
\myitems{
    \item If $|P| = 1$, our algorithm probes the only element $p \in P$, sets $Z = P$, and assigns $\weight(p) = 1$.
    \item If $|P| \ge 2$, our algorithm first produces the sets $P_\alpha$, $P_\middle$, and $P_\rest$ in \eqref{eqn:coreset:P_alpha}-\eqref{eqn:coreset:P_rest} using the function $G_1$ built in Section~\ref{sec:coreset:1d-alg}. If $P_\rest \ne \emptyset$, it adds to $Z$ the sample set $S_2$ described in Section~\ref{sec:coreset:1d-alg} and assigns $\weight(p) = |P_\rest| / |S_2|$ for each $p \in S_2$. If $P_\alpha \ne \emptyset$, it adds to $Z$ the sample set $S_\alpha$ described in Section~\ref{sec:coreset:1d-alg} and assigns $\weight(p) = |P_\alpha| / |S_\alpha|$ for each $p \in S_\alpha$. The recursion on $P_\middle$ returns a coreset $Z_\middle \subseteq P_\middle$, which is also included in $Z$.
}
}

The following pseudocode summarizes the above steps.
\rev{
\mytab{
    \> {\bf algorithm \ttt{BuildCoreset}} $(P)$ \\
    \> 1. \> $Z = \emptyset$ \\
    \> 2. \> {\bf if} $|P| = 1$ {\bf then} probe the (only) element $p \in P$, and set $Z = P$ with $\weight(p) = 1$ \\
    \> \> {\bf else} \\
    \> 3. \>\> probe the sample set $S_1$ described in Section~\ref{sec:coreset:1d-alg} \\
    \>\>\> /* this defines $G_1$ (see \eqref{eqn:coreset:G1-constructed}), which in turn defines $P_\alpha, P_\middle, P_\rest$ (see \eqref{eqn:coreset:P_alpha}-\eqref{eqn:coreset:P_rest}) */\\
    \> 4. \>\> {\bf if} $P_\rest \ne \emptyset$ {\bf then} \\
    \> 5.\>\>\> probe the sample set $S_2$ described in Section~\ref{sec:coreset:1d-alg} /* this defines $G_2$ (see \eqref{eqn:coreset:G2-constructed}) */ \\
    \> 6. \>\>\> add $S_2$ to $Z$ with $\weight(p) = |P_\rest|/|S_2|$ for each $p \in S_2$ \\
    \> 7. \>\> {\bf if} $P_\alpha \ne \emptyset$ {\bf then} \\
    \> 8. \>\>\> probe the sample set $S_\alpha$ described in Section~\ref{sec:coreset:1d-alg} /* this defines $F_\alpha$ (see \eqref{eqn:coreset:F_alpha-constructed}) */ \\
    \> 9. \>\>\> add $S_\alpha$ to $Z$ with $\weight(p) = |P_\alpha|/|S_\alpha|$ for each $p \in S_\alpha$ \\
    \> 10. \>\> $Z_\middle =$ \ttt{Build-Coreset} $(P_\middle)$;
    add $Z_\middle$ to $Z$ \\
    \> 11. {\bf return} $Z$
}
}

\noindent By the discussion in Section~\ref{sec:coreset:1d-alg}, \ttt{BuildCoreset} returns a set $Z$ of $O(\fr{\log n}{\eps^2} \cdot \log \fr{n}{\delta})$ elements. \rev{It is rudimentary to verify that for every $h \in \mono$, we have
\myeqn{
    \werr_Z(h) = F(h) \nn
}
where $F$ is given in \eqref{eqn:coreset:F-for-n-equas-1} if $|P| = 1$ or in \eqref{eqn:coreset:F-case2} otherwise. This establishes Theorem~\ref{thm:coreset} for $d = 1$.}

\subsection{Arbitrary Dimensions} \label{sec:coreset:any-d}

This subsection proves Theorem~\ref{thm:coreset} for any $d \ge 2$. As before, denote by $n$ and $w$ the size and width of the input $P$, respectively. We start by computing a chain decomposition of $P$ with $w$ chains: $C_1, C_2, ..., C_w$. Such a decomposition can be computed in time polynomial in $d$ and $n$ (see \cite{tw21}) without any probing.

\vgap

For every $i \in [w]$, we compute a subset $Z_i \subseteq C_i$ where every element $p \in Z_i$ has its label revealed and carries a weight $\weight(p) > 0$. The set $Z_i$ ensures
\myeqn{
    \err_{C_i}(h) \cdot \left(1-\fr{\eps}{4}\right) - \Delta_i
    \le
    \werr_{Z_i}(h)
    \le
    \err_{C_i}(h) \cdot \left(1-\fr{\eps}{4}\right) + \Delta_i
    \label{eqn:md-chain-Sigma-guarantee}
}
for every $h \in \mono$, where $\Delta_i$ is a possibly unknown value. Then, we obtain
\myeqn{
    Z &=& \bigcup_{i=1}^w Z_i. \label{eqn:coreset:md-Z}
}
For every $h \in \mono$, it holds that
\myeqn{
    \err_P(h) \cdot \left(1-\fr{\eps}{4}\right) - \Delta
    \le
    \werr_{Z}(h)
    \le
    \err_P(h) \cdot \left(1-\fr{\eps}{4}\right) + \Delta
    \label{eqn:coreset:md-Z-guarantee}
}
where
\myeqn{
    \Delta = \sum_{i=1}^w \Delta_i. \nn
}

Finding $Z_i$ for an $i \in [w]$ is a 1D problem. To explain, let us sort the elements of $C_i$ in ``ascending'' order (i.e., if $p$ precedes $q$ in the ordering, then $p \dombyeq q$). A monotone classifier $h$ maps a prefix of the ordering to $-1$; hence, as far as $C_i$ is concerned, we can regard $h$ as a 1D classifier of the form \eqref{eqn:coreset:1d-classifier}. As Theorem~\ref{thm:coreset} holds for $d = 1$, we can apply it to produce the desired $Z_i$ with probability at least $1 - \fr{\delta}{w}$ by probing $O(\fr{1}{\eps^2} \cdot \Log |C_i| \cdot \log (wn/\delta)) = O(\fr{\log (n/\delta)}{\eps^2} \Log |C_i|)$ elements from $C_i$. This $Z_i$ has size $O(\fr{\log (n/\delta)}{\eps^2} \Log |C_i|)$.

\vgap

Therefore, with probability at least $1 - 1/\delta$, we can produce $Z_1, ..., Z_w$ with
\myeqn{
    O\left( \fr{\log (n/\delta)}{\eps^2} \sum_{i=1}^w \log |C_i| \right)
    &=&
    O\left(\fr{\log (n/\delta)}{\eps^2} \cdot w \Log \fr{n}{w} \right) \nn
}
probes in total. The same bound also applies to the size of $Z$ in \eqref{eqn:coreset:md-Z}.

\section{Optimal Monotone Classification Needs ${\Omega(n)}$ Probes} \label{sec:lb-eps0}

This section focuses on Problem 1 under $\eps = 0$; namely, the objective is to find an optimal monotone classifier. A naive solution is to probe everything in the input $P$. We  prove that this is already the best approach up to a constant factor:

\begin{theorem} \label{thm:lb-eps0}
    For Problem 1, any algorithm promising to find an optimal classifier with probability \rev{greater than} $2/3$ must probe $\Omega(n)$ elements in expectation, where $n$ is the size of the input $P$. This is true even if the dimensionality $d$ is 1, and the algorithm knows the optimal monotone error $k^*$ of $P$.
\end{theorem}

The rest of the section serves as a proof of the theorem. Assume, w.l.o.g., that $n$ is an even number. We construct a family $\bbP$ of $n$ one-dimensional inputs. Every input of $\bbP$ has $n$ elements positioned at values 1, 2, ..., $n$. Various inputs differ in their label assignments. Specifically, each $i \in [n/2]$ defines two inputs in $\bbP$:
\myitems{
    \item $P_{-1}(i)$, where every odd (resp., even) element has label 1 (resp., $-1$), with $2i-1$ being the only exception, which is assigned label $-1$ instead. \rev{We refer to $P_{-1}(i)$ as a {\em $(-1)$-input}.}

    \item $P_{1}(i)$, where every odd (resp., even) element has label 1 (resp., $-1$), with $2i$ being the only exception, which is assigned label 1 instead. \rev{We refer to $P_1(i)$ as a {\em $1$-input}.}
}

The constructed family $\bbP = \set{P_{-1}(1), P_{-1}(2), ..., P_{-1}(n/2), P_{1}(1), P_{1}(2), ...,$ $P_{1}(n/2)}$ can be understood in an alternative manner. Group the elements $1, 2, ..., n$ into $n/2$ pairs $(1, 2), (3, 4),$ $..., (n-1, n)$. In a {{\em normal pair}} $(x-1, x)$, elements $x-1$ and $x$ carry labels 1 and $-1$, respectively. Each input $P \in \bbP$ contains one {{\em anomaly pair}} $(x-1, x)$: if $P$ is a $(-1)$-input, both $x-1$ and $x$ have label $-1$; otherwise, they have label 1.

\vgap

The optimal monotone error $k^*$ is $n/2 - 1$ for each input $P \in \bbP$. Indeed, every monotone classifier must misclassify at least one element in each normal pair. On the other hand, the error $n/2-1$ is attainable by mapping all the elements to 1 for a $1$-input or $-1$ for a $(-1)$-input.

\vgap

An algorithm $\A$ for Problem 1 {{\em errs}} on an input $P \in \bbP$ if it fails to find an optimal classifier for $P$. Denote by $\cost_P(\A)$ the number of probes performed by $\A$ when executed on $P$; this is a random variable if $\A$ is randomized. Define
\myeqn{
    \totalerr(\A) &=& \sum_{P \in \bbP} \Pr[\textrm{$\A$ errs on $P$}] \nn \\
    \totalcost(\A) &=& \sum_{P \in \bbP} \cost_P(\A). \nn
}
If $\A$ deterministic, $\Pr[\textrm{$\A$ errs on $P$}]$ is either 0 or 1 for each $P \in \bbP$. In Section~\ref{sec:lb-eps0:help}, we prove:

\begin{lemma} \label{lmm:lb-eps0:det}
    Fix any deterministic algorithm $\A_\det$ and an arbitrary non-negative constant $c < 1$. For $\rev{n \ge 4}$, if $\totalerr(\A_\det)$ $\le$ $cn/2$, then $\rev{\totalcost(\A_\det) \ge \fr{n^2}{4} (1-c^2)}$.
\end{lemma}

We now utilize the lemma to prove a hardness result for randomized algorithms.

\begin{corollary} \label{crl:lb-eps0:rand}
    When $n \ge 4$, the following holds for any randomized algorithm $\A$: if $\totalerr(\A) < n/3$, then $\expt[\totalcost(\A)] = \Omega(n^2)$.
\end{corollary}

\begin{proof}
    A randomized algorithm becomes deterministic when all the random bits are fixed. Hence, we can treat $\A$ as a distribution over a family $\bbA$ of deterministic algorithms, each sampled possibly with a different probability. We say that an algorithm $\A_\det \in \bbA$ {\em accurate} if $\totalerr(\A_\det) \le (2/5) n$. Define $\bbA_\mit{acc}$ as the set of accurate algorithms in $\bbA$.

    \vgap

    We argue that $\Pr[\A \in \bbA_\mit{acc}] > 1/6$. If $\Pr[\A \notin \bbA_\mit{acc}] \ge 5/6$, then
    \myeqn{
        \totalerr(\A) &=& \sum_{\A_\det \in \bbA} \totalerr(\A_\det) \cdot \Pr[\A = \A_\det] \nn \\
        &\ge&
        \sum_{\rev{\A_\det \notin \bbA_\mit{acc}}} \totalerr(\A_\det) \cdot \Pr[\A = \A_\det] \nn \\
        &\ge&
        \fr{2n}{5}\sum_{\rev{\A_\det \notin \bbA_\mit{acc}}} \Pr[\A = \A_\det] \nn \\
        &=& \fr{2n}{5} \cdot \Pr[\rev{\A \notin \A_\mit{acc}}] 
        \ge
        \fr{2n}{5} \cdot \fr{5}{6} = n/3 \nn
    }
    contradicting the condition $\totalerr(\A) < n/3$.

    \vgap

    By Lemma~\ref{lmm:lb-eps0:det}, every accurate $\rev{\bbA_\det \in \A_\det}$ satisfies $\totalcost(A_\det) = \Omega(n^2)$. We thus have $\expt[\totalcost (\A)] \ge \Omega(n^2) \cdot \Pr[\A \in \bbA_\mit{acc}] = \Omega(n^2)$.
\end{proof}

The corollary implies Theorem~\ref{thm:lb-eps0}. Indeed, if $\A$ guarantees finding an optimal classifier with probability \rev{greater than} $2/3$ on any input, then $\totalerr(\A) < |\bbP|/3 = n/3$. By Corollary~\ref{crl:lb-eps0:rand}, $\expt[\totalcost(\A)] = \Omega(n^2)$ when $n \ge 4$. This means that the expected cost of $\A$ is $\Omega(n)$ on at least one input in $\bbP$ since $\bbP$ has $n$ inputs.

\subsection{Proof of Lemma~\ref{lmm:lb-eps0:det}} \label{sec:lb-eps0:help}

We start with a property of the family $\bbP$.

\begin{proposition} \label{prop:lb-eps0:bbP-property}
    For any $i \in [n/2]$, no monotone classifier can be optimal for both $P_{-1}(i)$ and $P_{1}(i)$.
\end{proposition}

\begin{proof}
    As mentioned, the optimal monotone error is $n/2 - 1$ for each input of $\bbP$. A 1D monotone classifier $h$ has the form in \eqref{eqn:coreset:1d-classifier}, which is parameterized by a value $\tau$; next, we denote the classifier as $h^\tau$. To argue that no $h^\tau$ is optimal for both $P_{-1}(i)$ and $P_{1}(i)$, we examine all possible scenarios.
    \myitems{
        \item Case $\tau < 2i - 1$: on $P_{-1}(i)$, $h^\tau$ misclassifies both $2i-1$ and $2i$ and has error $n/2+1$.

        \vgap

        \item Case $\tau = 2i - 1$: on $P_{-1}(i)$, $h^\tau$ misclassifies $2i$ and has error $n/2$.

        \vgap

        \item Case $\tau \ge 2i$: on $P_{1}(i)$, $h^\tau$ misclassifies both $2i-1$ and $2i$ and has error $n/2+1$.
    }
    This completes the proof.
\end{proof}

To prove Lemma~\ref{lmm:lb-eps0:det}, we strengthen $\A_\det$ by giving it certain ``free'' labels. Every time $\A_\det$ probes an element of some pair $(2i-1, 2i)$, where $i \in [n/2]$, we reveal the label for the other element (of the pair) voluntarily. Henceforth, $\A_\det$ is said to ``probe pair $i$'' if $\A_\det$ probes either $2i-1$ or $2i$. If Lemma~\ref{lmm:lb-eps0:det} holds even on such an ``empowered'' $\A_\det$, it must hold on the original $\A_\det$ because an empowered algorithm can choose to ignore the free information.

\vgap

We consider, w.l.o.g., that $\A_\det$ terminates immediately after probing an anomaly pair --- identifying the pair enables $\A_\det$ to output an optimal classifier because the labels in normal pairs are fixed. Thus, we can model $\A_\det$ as a procedure probing a fixed sequence: pair $x_1$, pair $x_2$, ..., pair $x_t$ up to some $t$ $\in$ $[0, n/2]$. For each $j \in [t-1]$, if $x_j$ is an anomaly, $\A_\det$ terminates; otherwise, it moves on to probe $x_{j+1}$. If all the $t$ pairs are probed but no anomaly is found, $\A_\det$ outputs a fixed classifier, denoted as $h_\det$.

\vgap

As $\A_\det$ never probes pair $i$ for
\myeqn{
    i \in \{1, 2, ..., n/2\} \setminus \{x_1, x_2, ..., x_t\} \label{eqn:1dlb-help-1}
}
its output must be $h_\det$ on both $P_{-1}(i)$ and $P_{1}(i)$. By Proposition~\ref{prop:lb-eps0:bbP-property}, $h_\det$ cannot be optimal for both $P_{-1}(i)$ and $P_{1}(i)$, which gives
\myeqn{
    \totalerr(\A_\det) &\ge& n/2 - t. \label{eqn:1dlb-help-2}
}
To analyze its cost, note that $\A_\det$ performs $t$ probes for $P_{-1}(i)$ and $P_{1}(i)$ of every $i$ satisfying \eqref{eqn:1dlb-help-1}, but $j \in [t]$ probes for $P_{-1}(x_j)$ and $P_{1}(x_j)$. Hence
\myeqn{
    \totalcost(\A_\det) = 2 t \cdot (n/2 - t) + 2 \sum_{j=1}^t j  
    = nt - t^2 \rev{\,+\,} t. \label{eqn:1dlb-help-3}
}

If $\totalerr(\A_\det)$ needs to be at most $cn/2$, then $t \ge \fr{n}{2} (1-c)$ by \eqref{eqn:1dlb-help-2}. On the other hand, for $\rev{t \in [\fr{n}{2} (1-c), \fr{n}{2}]}$, we have
\rev{
$
    \eqref{eqn:1dlb-help-3} \ge nt - t^2 \ge \fr{n^2}{4} (1-c^2).
$
This completes the proof of Lemma~\ref{lmm:lb-eps0:det}.}

\section{A Lower Bound for Constant Approximation Ratios} \label{sec:lb-eps-const}

We now proceed to study the hardness of approximation. The main result of this section is:

\begin{theorem} \label{thm:lb-eps-const}
    Let $n'$, $w'$, $k$, and $c$ be arbitrary integers satisfying $n' \ge 2, w' \ge 1, k \ge 0$, $c \ge 1$, and $n'$ is a multiple of $w'$. Set
    \myeqn{
        n &=& n' + 2k + 2ck n' \label{eqn:lb-eps-const:n} \\
        w &=& w' + \mathbbm{1}_{k \ge 1}. \label{eqn:lb-eps-const:w}
    }
    For Problem 1, there is a family $\bbP$ of inputs with size $n$, width $w$, and optimal monotone error $k^* = k$ such that any randomized algorithm, which guarantees an expected error at most $c \cdot k^*$, must entail an expected cost of $\Omega(w' \Log \fr{n'}{w'})$ on at least one input of $\bbP$, where $\Omega(.)$ hides a constant that is not dependent on $n'$, $w'$, $k$, and $c$. The claim holds even if the algorithm knows $k^*$.
\end{theorem}

The theorem is particularly useful when the approximation ratio $c$ is a constant. When $k = 0$ (realizable), the theorem gives a lower bound $\Omega(w \Log \fr{n}{w})$. For $k \ge 1$ (non-realizable), we always have $n' + 2k \le n /2$, because of which
\myeqn{
    \fr{n'}{w'}
    =
    \fr{n - (n' + 2k)}{2c k w'}
    \ge
    \fr{n}{4c kw}. \nn
}
Theorem~\ref{thm:lb-eps-const} thus implies a lower bound $\Omega(w \Log \fr{n}{k^* w})$ when $w$ is sufficiently large. We first prove the theorem for $k = 0$ in Section~\ref{sec:lb-eps-const:realizable} and then for $k \ge 1$ in Section~\ref{sec:lb-eps-const:nonrealizable}.

\subsection{The Realizable Case} \label{sec:lb-eps-const:realizable}

We use the term {{\em box}} to refer to an axis-parallel rectangle with a positive area in $\real^2$. The {{\em main diagonal}} of the box is the segment connecting its bottom-left and top-right corners. We say that two boxes $B_1$ and $B_2$ are {{\em independent}} if no point in $B_1$ dominates any point in $B_2$ and vice versa.

\vgap

For $k^* = 0$, an algorithm that guarantees an expected error at most $c k^* = 0$ must always find an optimal classifier. To prove Theorem~\ref{thm:lb-eps-const} for $k = 0$, we construct hard inputs as follows. Let $B_1, B_2, ..., B_{w'}$ be arbitrary mutually independent boxes. For each $i \in [w']$, place $n'/w'$ points on the main diagonal of $B_i$, making sure that they are at distinct locations and no point lies at a corner of $B_i$. This yields a set $P$ of $n'$ points with  width $w'$; see Figure~\ref{fig:lb-eps-constant:realizable}. Label assignment is done for each box independently, subject to the constraint that $P$ is monotone. In each box, there are $1 + \fr{n'}{w'}$ assignments: for each $i \in [0, n'/w']$, assign label $-1$ to the $i$ lowest points in the box and 1 to the rest. This gives a family $\bbP_{n',w'}$ of $(1 + n'/w')^{w'}$ labeled point sets.

\begin{figure}
    \centering
    \includegraphics[height=40mm]{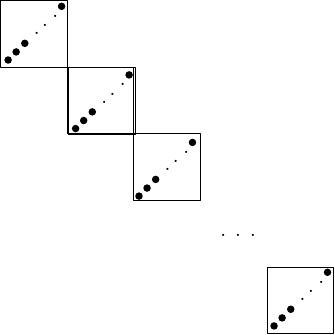}
    \figcapup
    \caption{A hard realizable input for Problem 1: $w$ boxes each with $n/w$ points}
    \label{fig:lb-eps-constant:realizable}
    \figcapdown
\end{figure}

\vgap

A deterministic algorithm $\A_\det$ is a binary decision tree $\T$. If $\A_\det$ is always correct for $k = 0$, it must distinguish all the inputs in $\bbP_{n',w'}$ by returning a different classifier for each input (no classifier is optimal for two inputs in $\bbP_{n',w'}$). The number of leaves in $\T$ is thus at least $(1 + n'/w')^{w'}$. The {{\em average cost}} of $\A_\det$ --- \rev{where the average is over} the inputs of $\bbP_{n',w'}$ --- equals the average depth of the leaves in $\T$. A binary tree with at least $(1 + n'/w')^{w'}$ leaves must have an average depth of $\Omega(w' \Log \fr{n'}{w'})$. Hence, $\A_\det$ must have average cost $\Omega(w \Log \fr{n}{w})$.

\vgap

By Yao's minimax theorem \cite{mp95}, any randomized algorithm that is always correct for $k = 0$ must entail an expected cost of $\Omega(w' \Log \fr{n'}{w'})$ on at least one input of $\bbP_{n',w'}$.

\subsection{The Non-Realizable Case} \label{sec:lb-eps-const:nonrealizable}

This subsection serves as a proof of Theorem~\ref{thm:lb-eps-const} for $k \ge 1$.

\extraspacing {\bf Algorithms with Guessing Power.} We first strengthen the power of deterministic algorithms. As before, such an algorithm $\A_\det$ is described by a binary decision tree $\T$ determined by the point locations in the input $P$. Different from the decision tree in Section~\ref{sec:intro:prob}, however, we allow two types of internal nodes:
\begin{itemize}
    \item {{\em Probe node}}. This is the (only) type of internal nodes allowed in Section~\ref{sec:intro:prob}.

    \item {{\em Guess node}}. At such a node, $\A_\det$ proposes a monotone classifier $h$ and asks an \rev{oracle} whether $\err_P(h)$ is at most a certain value \rev{$t$. Both $h$ and $t$ are fixed at this node.} On a ``yes'' answer from the \rev{oracle}, $\A_\det$ descends to the left child, which must be a leaf returning $h$. On a ``no'' answer, $\A_\det$ branches right and continues.
\end{itemize}
We charge one unit of cost \rev{for every node: probe or guess}. A randomized algorithm is still modeled as a function that maps a random-bit sequence to a deterministic algorithm. A lower bound on such empowered algorithms must also hold on algorithms that use probe nodes only.

\vgap

The argument in Section~\ref{sec:lb-eps-const:realizable} has  proved that any deterministic algorithm in the form of a binary decision tree must have an average cost of $\Omega(w' \Log\fr{n'}{w'})$ over the inputs of $\bbP_{n',w'}$. Hence, the lower bound applies to  deterministic algorithms with guess nodes. By Yao's minimax theorem, any randomized algorithm with guess nodes must incur $\Omega(w' \Log\fr{n'}{w'})$ expected cost on at least one input of $\bbP(n', w')$ if it always returns an optimal classifier.

\extraspacing {\bf A Las Vegas Lower Bound.} Let $n', k', k, c, n$, and $w$ be as stated in Theorem~\ref{thm:lb-eps-const}. Denote by $\A$ a randomized algorithm (with guessing power) that, when executed on an input having size $n$, width $w$, and optimal monotone error $k^* = k$, guarantees
\myitems{
    \item returning a monotone classifier whose error on the input is at most $ck^*$, and
    \item an expected cost at most $J_\mit{LV}$.
}
We show that $J_\mit{LV} = \Omega(w' \Log \fr{n'}{w'})$ \rev{by using $\A$ to process the inputs from $\bbP(n', w')$}.

\begin{figure*}
    \centering
    \begin{tabular}{cc}
        \includegraphics[height=40mm]{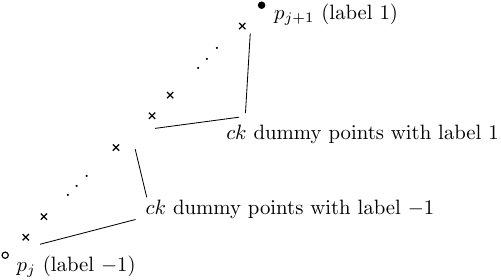} &
        \includegraphics[height=40mm]{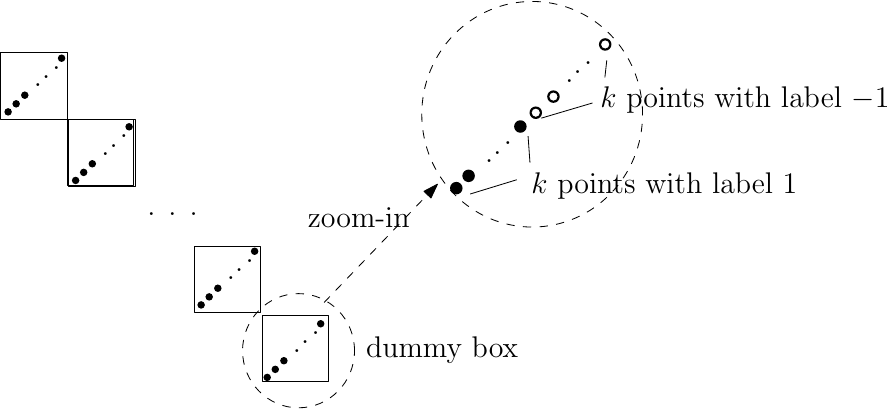} \\
        (a) Dummy points in a non-dummy box &
        (b) Dummy points in the dummy box
    \end{tabular}
    \figcapup
    \caption{Adding dummy points for a Las Vegas lower bound}
    \label{fig::probOne-lb}
    \figcapdown
\end{figure*}

\vgap

Given an input $P' \in \bbP_{n',w'}$, we construct a set $P$ of labeled points as follows. First, set $P = P'$. Recall that the points of $P'$ are inside $w'$ boxes $B_1, B_2, ..., B_{w'}$, each having $n'/w'$ points. For each $i \in [w']$, let $p_1, p_2, ..., p_{n'/w'}$ be the points of $P$ in $B_i$, sorted in ascending order of y-coordinate. For each $j \in [n'/w' - 1]$, place $2ck$ dummy points on the main diagonal of $B_i$ between $p_j$ and $p_{j+1}$. Assign labels to these dummy points as follows:
\begin{itemize}
    \item If $\lab(p_j) = \lab(p_{j+1})$, set the labels of all $2ck$ dummy points to $\lab(p_j)$.

    \vgap

    \item Otherwise, we must have $\lab(p_j) = -1$ and $\lab(p_{j+1}) = 1$ (because $P'$ is monotone); set the labels of the $ck$ lowest dummy points to $-1$ and the labels of the other dummy points to $1$; see Figure~\ref{fig::probOne-lb}a.
\end{itemize}
Furthermore, add $ck$ dummy points to $P$ between the bottom-left corner of $B_i$ and $p_1$. Set their labels to $-1$ if $\lab(p_1) = -1$, or $1$ otherwise. Symmetrically, add to $P$ another $ck$ dummy points between $p_{n'/w'}$ and the top-right corner of $B_i$. Set their labels to $1$ if $\lab(p_{n'/w'}) = 1$, or $-1$ otherwise. Finally, create a dummy box that is independent from all of $B_1, ..., B_{n'/w'}$. Add $2k$ points to $P$ on the main diagonal of this box, setting the labels of the $k$ lowest (resp., highest) points to 1 (resp., $-1$); see Figure~\ref{fig::probOne-lb}b. This finishes the construction of $P$.

\vgap

The set $P$ has $n = n' + 2k + 2ck n'$ points (all at distinct locations) and width $w = w' + 1$. Furthermore, the optimal monotone error $k^*$ of $P$ is $k$ because (i) any monotone classifier must misclassify at least $k$ points in the dummy box, and (ii) error $k$ is attainable by the classifier that classifies all the points in $B_1, ..., B_{n'/w'}$ correctly and maps all the points in the dummy box to 1.

\vgap

Let us apply the given algorithm $\A$ on $P$ and obtain its output classifier $h$. We argue that $h$ must correctly classify every non-dummy point $p \in P$ (every such $p$ originated from $P'$). Otherwise, suppose that $\lab(p) = -1$ but $h(p) = 1$ for some non-dummy $p \in P$. By our construction, $p$ is dominated by at least $ck$ points of label $-1$. As $h$ maps all those $c k$ points to 1, we know $\err_P(h) \ge ck + 1$, contradicting the fact that $\A$ guarantees an error at most $c k^*$. A symmetric argument rules out the possibility that $\lab(p) = 1$ but $h(p) = -1$.  We can thus return $h$ as an optimal classifier for $P'$. It follows from our earlier lower bound on $\bbP(n',w')$ that $J_\mit{LV}$ must be $\Omega(w' \Log \fr{n'}{w'})$.

\extraspacing {\bf A Monte-Carlo Lower Bound.} Again, let $n', k', k, c, n$, and $w$ be as stated in Theorem~\ref{thm:lb-eps-const}. Let $\A$ be a randomized algorithm (with guessing power) that, when executed on an input $P$ with size $n$, width $w$, and optimal monotone error $k^* = k$, always guarantees
\myitems{
    \item an expected error at most $ck$ on $P$ and
    \item an expected cost at most $J_\mit{MC}$.
}
Next, we show $J_\mit{MC} = \Omega(w' \Log \fr{n'}{w'})$, which will complete the proof of Theorem~\ref{thm:lb-eps-const}.

\vgap

With probability at least 1/2, the algorithm $\A$ must (i) return a monotone classifier whose error on $P$ is at most $4ck$ and (ii) probe at most $4 J_\mit{MC}$ points. Otherwise, one of the following two events must occur with probability at least 1/4:
\myitems{
    \item $\A$ outputs a classifier whose error on $P$ is over $4ck^*$, or
    \item $\A$ probes more than $4 J_\mit{MC}$ points of $P$.
}
However, this means that $\A$ either has an expected error higher than $ck^*$ or incurs an expected cost higher than $J_\mit{MC}$, contradicting its guarantees.

\vgap

We deploy $\A$ as a black box to design a randomized algorithm that probes $O(J_{MC})$ points in expectation and {\em always} returns a monotone classifier with an error at most $4ck^*$ on $P$. For this purpose, run $\A$ until either it returns a monotone classifier $h$ or has probed $4J_{MC}$ points. In the former situation, we ask the \rev{oracle} whether $\err_P(h) \le 4ck^*$. If so, return $h$. In all other situations (i.e., the \rev{oracle} answers ``no'' or $\A$ does not terminate after $4J_{MC}$ probes), we declare ``failure'' and start all over again. After having failed $\lc \log_2 n \rc$ times, we simply probe the entire $P$ and return an optimal monotone classifier. Since each time we fail with probability at most $1/2$, the expected probing cost is bounded by
\begin{eqnarray}
    \left(\sum_{i=0}^{\lc \log_2 n \rc}  4 J_\mit{MC} \cdot (i+1) \left(\fr{1}{2} \right)^i \right)  + \fr{1}{n} \cdot n 
    = O(J_{MC}) \nn
\end{eqnarray}
noticing that the probability of failing $\lc \log_2 n \rc$ times is at most $1/n$. By our earlier Las Vegas lower bound, we conclude that $J_{MC} = \Omega(w' \Log \fr{n'}{w'})$.

\section{A Lower Bound for Arbitrary Approximation Ratios} \label{sec:lb-eps-any}

We now continue our study on the hardness of approximation in the regime where $\eps$ can be arbitrarily small. The main result in this section is:

\begin{theorem} \label{thm:lb-eps-any}
    Let $\eps$ be an arbitrary value satisfying $0 < \eps \le 1/10$. Fix any integers $n \ge 2$ and $w \ge 1$ such that $n$ is a multiple of $w$, and $n \ge \fr{w}{\eps^2} \ln n$. Suppose that $\A$ is an algorithm for Problem 1 under $d = 2$ that guarantees an expected error of at most $(1+\eps) k^*$ on every input of size $n$, where $k^*$ is the optimal monotone error. Then, the expected cost of $\A$ must be $\Omega(w / \eps^2)$ on at least one input with width $w$, where $\Omega(.)$ hides a constant that does not depend on $\eps$, $n$, $w$, and $k^*$.
\end{theorem}

The rest of the section serves as a proof of the theorem. Define
\myeqn{
    \gamma &=& 9 \eps \label{eqn:lb-eps-any:gamma} \\
    \mu_1 &=& (1-\gamma)/2 \label{eqn:lb-eps-any:mu1} \\
    \mu_2 &=& (1+\gamma)/2 \label{eqn:lb-eps-any:mu2} \\
    M &=& (1-\gamma^2) / (9\gamma^2).
        \label{eqn:lb-eps-any:M}
}
Because $\gamma=9\eps$ and $\eps \le 1/10$, we have
\myeqn{
    \fr{wM}{8}
    =
    \fr{1-81\eps^2}{5832} \cdot \fr{w}{\eps^2}
    >
    \fr{1}{31000} \fr{w}{\eps^2}.
    \label{eqn:lb-eps-any:wm/6-is-big-Omega}
}

Suppose, for contradiction, that $\A$ has expected cost at most $wM/8$ on every input of size $n$ and width $w$:
\myeqn{
    \A = \text{an algorithm with expected cost at most $wM/8$ on every input of size $n$ and width $w$.}
    \label{eqn:lb-eps-any:A}
}

Let $x_1, x_2, ..., x_w$ be $w$ distinct locations in $\real^2$ such that no location dominates another. Henceforth, we fix $P$ to be a multiset of $n$ points where
\myitems{
    \item exactly $n/w$ points are placed at location $x_i$ for each $i \in [w]$;
    \item every element of $P$ has a distinct ID.
}
The elements in $P$ do not carry labels yet. The random process below stochastically generates their labels and measures the cost and error of $\A$ over the resulting $P$:

\mytab{
    \> \bf{\ttt{RP-1}} \\
    \> 1. \> $\bm{\mu} =$ a $w$-dimensional vector sampled from $\set{\mu_1, \mu_2}^{w}$ uniformly at random \\
    \> 2. \> {\bf for} every element $p \in P$ {\bf do} /* suppose that $p$ is at location $x_i$ */ \\
    \> 3. \>\> assign $p$ label 1 with probability $\bm{\mu}[i]$ or $-1$ with probability $1 - \bm{\mu}[i]$ \\
    \> 4. \> $h_\A =$ the monotone classifier output by algorithm $\A$ when executed on $P$; \\
    \>\> $\Lambda_1 =$ the number of probes $\A$ performed \\
    \> 5. \> $R_1 = \err_P(h_\A)$ \\
    \> 6. \> {\bf return} $(\Lambda_1, R_1)$
}

\noindent The optimal monotone error $k^*$ is a random variable determined by the labeled multiset $P$. By \eqref{eqn:lb-eps-any:A}, we have $\expt[\Lambda_1] \le wM/8$. Our objective is to show that the expected error of $\A$ is too large. We do so by relating \ttt{RP-1} to another random process:

\vgap

\mytab{
    \> \bf{\ttt{RP-2}} \\
    \> 1. \> $\bm{\mu} =$ a $w$-dimensional vector sampled from $\set{\mu_1, \mu_2}^{w}$ uniformly at random \\
    \>\> /* now run $\A$ on $P$ */ \\
    \> 2. \> {\bf while} algorithm $\A$ still needs to perform a probe {\bf do} \\
    \> 3. \>\> $p =$ the element probed by $\A$ (identified by ID) /* suppose that $p$ is at location $x_i$ */ \\
    \> 4. \>\> assign $p$ label 1 with probability $\bm{\mu}[i]$ or $-1$ with probability $1 - \bm{\mu}[i]$ \\
    \> 5. \> $h_\A =$ the monotone classifier output by $\A$; \\
    \>\> $\Lambda_2 =$ the number of probes $\A$ performed \\
    \> 6. \> {\bf for} every element $q \in P$ that has not been probed by $\A$ {\bf do} /* suppose that $q$ is at $x_i$ */ \\
    \> 7. \>\> assign $q$ label 1 with probability $\bm{\mu}[i]$ or $-1$ with probability $1 - \bm{\mu}[i]$ \\
    \> 8. \> $R_2 = \err_P(h_\A)$ \\
    \> 9. \> {\bf return} $(\Lambda_2, R_2)$
}

\begin{lemma} \label{lmm:lb-eps-any:U1=U2}
    $\expt [\Lambda_1] = \expt [\Lambda_2]$ and $\expt [R_1] = \expt [R_2]$.
\end{lemma}

\begin{proof}
    We prove only $\expt [R_1] = \expt [R_2]$ because an analogous (and simpler) argument shows $\expt [\Lambda_1] = \expt [\Lambda_2]$. Let us first consider that $\A$ is a deterministic algorithm, i.e., a binary decision tree $\T$. Recall that each internal node of $\T$ is associated with an element (identified by ID) in $P$ that should be probed when $\A$ is at this node. Each leaf of $\T$ is associated with a classifier that should be returned when $\A$ is at this node. For each leaf node $v$, denote by $\pi_v$ the path from the root of $\T$ to $v$.

    \vgap

    We have for each $j \in \set{1, 2}$:
    \myeqn{
        \expt[R_j] &=& \sum_{\text{leaf $v$ of $\T$}} \Pr[\text{$\A$ finishes at $v$ in \ttt{RP-$j$}}]
        \cdot
        \expt[R_j \mid \text{$\A$ finishes at $v$ in \ttt{RP-$j$}}]. \nn
    }

    Let us concentrate on an arbitrary leaf $v$. Let $p_1, p_2, ..., p_t$ be the elements associated with the internal nodes of $\pi_v$ in the top-down order. The algorithm $\A$ arrives at $v$ if and only if each element $p_i$ ($i \in [t]$) takes a specific label, denoted as $l_i$. The probability of the event ``$\lab(p_i)=l_i$ for all $i \in [t]$'' is identical in \ttt{RP-1} and \ttt{RP-2}. Hence, $\A$ has the same probability of reaching $v$ in each random process. Conditioned on the aforementioned event, the classifier $h_\A$ is fixed, while the labels of the elements in $P \setm \set{p_1, ..., p_t}$ have the same distribution in the two random processes. Therefore, the conditional expectations of $R_1$ and $R_2$ are identical. It follows that $\expt[R_1] = \expt[R_2]$.

    \vgap

    As a randomized algorithm is a distribution over a family of deterministic algorithms, the equality also holds for a randomized $\A$.
\end{proof}

Because $\expt[\Lambda_1] \le wM/8$, Lemma~\ref{lmm:lb-eps-any:U1=U2} gives
\myeqn{
    \expt[\Lambda_2] \le wM/8. \label{eqn:lb-eps-any:L2}
}
The labeled multiset $P$, and hence $k^*$, has the same distribution in \ttt{RP-1} and \ttt{RP-2}.
The next subsection proves:

\begin{lemma} \label{lmm:lb-eps-any:U2}
    When \eqref{eqn:lb-eps-any:L2} holds, $n \ge 2$, $\eps \le 1/10$, and $n \ge \fr{w}{\eps^2} \ln n$, we must have
    \myeqn{
        \expt[R_2] > (1+\eps) \expt[k^*]. \nn
    }
\end{lemma}

We can now complete the proof of Theorem~\ref{thm:lb-eps-any}. If $\A$ guarantees an expected error of at most $(1+\eps)k^*$ on every labeled input, then, after conditioning on the labeled multiset $P$ generated by \ttt{RP-1}, we have
\myeqn{
    \expt[R_1 \mid P] \le (1+\eps) k^*. \nn
}
Taking expectation over $P$ gives $\expt[R_1] \le (1+\eps)\expt[k^*]$. This contradicts Lemma~\ref{lmm:lb-eps-any:U1=U2} and Lemma~\ref{lmm:lb-eps-any:U2}. Hence, $\A$ must have expected cost greater than $wM/8$ on at least one input. The theorem now follows from \eqref{eqn:lb-eps-any:wm/6-is-big-Omega}.

\subsection{Proof of Lemma~\ref{lmm:lb-eps-any:U2}}

Our proof is inspired by a lower bound argument in \cite{bdl09} for non-realizable active learning. We need the following expected-sample testing bound, whose proof is deferred to the end of this subsection.

\begin{lemma} \label{lmm:lb-eps-any:distinguish}
    Define $\mu$ to be a random variable that equals $\mu_1$ or $\mu_2$ --- see \eqref{eqn:lb-eps-any:mu1} and \eqref{eqn:lb-eps-any:mu2} --- each with probability $1/2$. Consider an algorithm that adaptively draws i.i.d.\ samples from $\set{-1,1}$, where each sample equals 1 with probability $\mu$, and eventually outputs a guess from $\set{\mu_1,\mu_2}$. Let $X$ be its number of samples. If $\expt[X] < M$, where the expectation is over $\mu$, the samples, and the internal randomness of the algorithm, then its guess is wrong with probability greater than $1/3$.
\end{lemma}

The subsequent discussion is carried out under the conditions stated in Lemma~\ref{lmm:lb-eps-any:U2}. For each $i \in [w]$, let $M_i$ be the number of points at $x_i$ probed by $\A$ in \ttt{RP-2}. We say that $i$ is {\em light} if $\expt[M_i] < M$ or {\em heavy} otherwise.

\begin{lemma} \label{lmm:lb-eps-any:at-least-w/2-light}
    At least $7w/8$ values of $i$ are light.
\end{lemma}

\begin{proof}
    Every heavy $i$ satisfies $\expt[M_i] \ge M$. Because
    \myeqn{
        \sum_{i=1}^{w} \expt[M_i]
        =
        \expt[\Lambda_2]
        \le
        \fr{wM}{8}, \nn
    }
    the number of heavy values is at most $w/8$. The lemma follows.
\end{proof}

Consider the classifier $h_\A$ output by $\A$ at Line 5 of \ttt{RP-2}. For each $i \in [w]$, set $K_i = 1$ if one of the following events occurs:
\myitems{
    \item $h_\A(x_i) = 1$ and $\bm{\mu}[i] = \mu_1$;
    \item $h_\A(x_i) = -1$ and $\bm{\mu}[i] = \mu_2$.
}
Otherwise, set $K_i = 0$.

\begin{lemma} \label{lmm:lb-eps-any:K_i-expt}
    If $i \in [w]$ is light, then $\expt[K_i] > 1/3$.
\end{lemma}

\begin{proof}
    We interpret $h_\A(x_i)$ as a guess about $\bm{\mu}[i]$: the value 1 represents the guess $\bm{\mu}[i]=\mu_2$, while the value $-1$ represents the guess $\bm{\mu}[i]=\mu_1$. Thus, $K_i=1$ if and only if the guess is wrong.

    \vgap

    Conditional on $\bm{\mu}[i]$, the labels revealed by the probes at $x_i$ are i.i.d.\ samples of the form considered in Lemma~\ref{lmm:lb-eps-any:distinguish}. Because $i$ is light, $\expt[M_i]<M$. Lemma~\ref{lmm:lb-eps-any:distinguish} gives $\Pr[K_i=1]>1/3$, proving the lemma.
\end{proof}

Recall that \ttt{RP-2} randomly chooses a vector $\bm{\mu}$ at Line 1. For each $i \in [w]$, define its {{\em good $i$-label}} to be
\myitems{
    \item $1$ if $\bm{\mu}[i] = \mu_2$;
    \item $-1$ otherwise.
}
The label in $\set{-1,1}$ different from the good $i$-label is called the {{\em bad $i$-label}}. At $x_i$, the expected number of points receiving the good $i$-label is $\fr{n}{w}(\fr{1}{2}+\fr{\gamma}{2})$. We say that the labeled multiset $P$ created by \ttt{RP-2} is {{\em intended}} if, for every $i \in [w]$, at least
\myeqn{
    \fr{n}{w} \left(\fr{1}{2} + \fr{\gamma}{4} \right) \nn
}
points at location $x_i$ receive the good $i$-label.

\begin{lemma} \label{lmm:lb-eps-any:Pr-intended}
    The probability for $P$ to be intended is at least $1 - 1/n^4$.
\end{lemma}

\begin{proof}
    Focus on an arbitrary $i \in [w]$, and let $B_i$ be the number of points at $x_i$ receiving the bad $i$-label. Then
    \myeqn{
        \expt[B_i]
        =
        \fr{n}{w}\left(\fr{1}{2}-\fr{\gamma}{2}\right). \nn
    }
    Set $t=\fr{\gamma}{2(1-\gamma)}$, for which
    \myeqn{
        (1+t)\expt[B_i]
        =
        \fr{n}{w}\left(\fr{1}{2}-\fr{\gamma}{4}\right). \nn
    }
    By Chernoff bound \eqref{eqn:chernoff2},
    \myeqn{
        \Pr\left[
            B_i \ge \fr{n}{w}\left(\fr{1}{2}-\fr{\gamma}{4}\right)
        \right]
        &\le&
        \exp\left(-\fr{t^2}{2+t}\expt[B_i]\right) \nn \\
        &\le&
        \exp\left(-\fr{\gamma^2 n}{16w}\right) \nn \\
        \explain{by \eqref{eqn:lb-eps-any:gamma}}
        &=&
        \exp\left(-\fr{81\eps^2 n}{16w}\right). \nn
    }
    This probability is at most $1/n^5$ because $n \ge \fr{w}{\eps^2}\ln n$. A union bound over the $w \le n$ locations shows that $P$ is intended with probability at least $1-1/n^4$.
\end{proof}

\begin{lemma} \label{lmm:lb-eps-any:sum-of-K_i-when-P-intended}
    \myeqn{
        \expt\left[
            \mathbbm{1}_{\text{$P$ intended}}
            \sum_{i=1}^{w} K_i
        \right]
        >
        w\left(\fr{7}{24}-\fr{1}{n^4}\right). \nn
    }
\end{lemma}

\begin{proof}
    Lemmas~\ref{lmm:lb-eps-any:at-least-w/2-light} and \ref{lmm:lb-eps-any:K_i-expt} imply
    \myeqn{
        \expt\left[\sum_{i=1}^{w}K_i\right]
        >
        \fr{7w}{8}\cdot\fr{1}{3}
        =
        \fr{7w}{24}. \nn
    }
    On the other hand, $\sum_{i=1}^{w}K_i \le w$, while Lemma~\ref{lmm:lb-eps-any:Pr-intended} gives $\Pr[\text{$P$ not intended}] \le 1/n^4$. Therefore,
    \myeqn{
        \expt\left[
            \mathbbm{1}_{\text{$P$ intended}}
            \sum_{i=1}^{w}K_i
        \right]
        &\ge&
        \expt\left[\sum_{i=1}^{w}K_i\right]
        -
        w \cdot \Pr[\text{$P$ not intended}] \nn \\
        &>&
        w \cdot \left(\fr{7}{24}-\fr{1}{n^4}\right) \nn
    }
    as claimed.
\end{proof}

For every labeled multiset $P$ supported at $x_1, ..., x_w$, a monotone classifier can choose the majority label independently at every location because the locations are mutually incomparable. Hence,
\myeqn{
    k^* \le n/2. \label{eqn:lb-eps-any:k-star}
}

Suppose that $P$ is intended. For each $i \in [w]$, let $I_i$ be the number of points at $x_i$ receiving the good $i$-label. An optimal classifier $h^*$ maps $x_i$ to the good $i$-label, and
\myeqn{
    I_i \ge \fr{n}{w}\left(\fr{1}{2}+\fr{\gamma}{4}\right). \nn
}
By the definition of $K_i$, the classifiers $h^*$ and $h_\A$ disagree at $x_i$ if and only if $K_i=1$. Such a disagreement causes $h_\A$ to make $2I_i-n/w$ more errors than $h^*$ at $x_i$. Consequently, whenever $P$ is intended,
\myeqn{
    R_2-k^*
    =
    \err_P(h_\A)-k^*
    =
    \sum_{i=1}^{w}K_i\left(2I_i-\fr{n}{w}\right)
    \ge
    \fr{n\gamma}{2w}\sum_{i=1}^{w}K_i.
    \label{eqn:lb-eps-any:err-minus-k-star}
}

Because $R_2 \ge k^*$ always, Lemma~\ref{lmm:lb-eps-any:sum-of-K_i-when-P-intended} and \eqref{eqn:lb-eps-any:err-minus-k-star} yield
\myeqn{
    \expt[R_2]-\expt[k^*]
    &=&
    \expt[R_2-k^*] \nn \\
    &\ge&
    \expt\left[
        \mathbbm{1}_{\text{$P$ intended}}(R_2-k^*)
    \right] \nn \\
    &\ge&
    \fr{n\gamma}{2w}
    \expt\left[
        \mathbbm{1}_{\text{$P$ intended}}
        \sum_{i=1}^{w}K_i
    \right] \nn \\
    &>&
    \fr{n\gamma}{2}
    \left(\fr{7}{24}-\fr{1}{n^4}\right) \nn \\
    \explain{by $n \ge 2$ and \eqref{eqn:lb-eps-any:gamma}}
    &\ge&
    \fr{99\eps n}{96} \nn \\
    &>&
    \fr{\eps n}{2} \nn \\
    \explain{by \eqref{eqn:lb-eps-any:k-star}}
    &\ge&
    \eps\expt[k^*]. \nn
}
It follows that $\expt[R_2]>(1+\eps)\expt[k^*]$, completing the proof of Lemma~\ref{lmm:lb-eps-any:U2}.

\extraspacing {\bf Proof of Lemma~\ref{lmm:lb-eps-any:distinguish}.}
    For any $p \in [0,1]$, let $\operatorname{Bern}(p)$ denote the Bernoulli distribution over $\set{-1,1}$ that assigns probability $p$ to 1 and probability $1-p$ to $-1$. Define
    \myeqn{
        d_{\mu_1,\mu_2}
        =
        D_{\mit{KL}}\left(\operatorname{Bern}(\mu_1) \Vert \operatorname{Bern}(\mu_2)\right)
        =
        \gamma \ln \fr{1+\gamma}{1-\gamma} \nn
    }
    where $D_{\mit{KL}}$ represents KL divergence. It is easy to verify
    \myeqn{
        d_{\mu_1,\mu_2}
        =
        D_{\mit{KL}}\left(\operatorname{Bern}(\mu_2) \Vert \operatorname{Bern}(\mu_1)\right)
        =
        \gamma \ln \fr{1+\gamma}{1-\gamma}.
        \label{eqn:lb-eps-any:distribution:d-u1u2}
    }

    Define a {\em transcript} as the random bits consumed by the algorithm and the sequence of samples drawn. We will view a transcript as a sequence: $\Sigma_0, (\textrm{sample 1}, \Sigma_1)$, $(\textrm{sample 2}, \Sigma_2)$, ..., $(\textrm{sample $X$}, \Sigma_X)$ where, for each $0 \le i \le X-1$, $\Sigma_i$ is the random bit sequence consumed by the algorithm between samples  $i$ and $i+1$, and $\Sigma_X$ is the random bit sequence consumed after the last sample. The output of the algorithm is a function of the transcript.

    \vgap

    Let $\TTT$ be the space of all possible transcripts. For $j \in \set{1,2}$, define the probability distribution $\Q_j$ over $\TTT$ by
    \myeqn{
        \Q_j(\TT)
        &=&
        \Pr[\text{the transcript belongs to $\TT$} \mid \mu=\mu_j]
        \qquad
        \text{for every $\TT \subseteq \TTT$}. \nn
    }
    For each $T \in \TT$, we abbreviate $\Q_j(\set{T})$ as $\Q_j(T)$.
    The total-variation distance of $\Q_1$ and $\Q_2$ is
    \myeqn{
        d_{\mit{TV}}(\Q_1,\Q_2)
        &=&
        \sup_{\text{events $\TT \subseteq \TTT$}}
        \left|\Q_1(\TT)-\Q_2(\TT)\right|. \label{eqn:lb-eps-any:distribution:tv-def}
    }

    \begin{lemma} \label{lmm:lb-eps-KL-Q1Q2}
        $D_{\mit{KL}}(\Q_1 \Vert \Q_2) = d_{\mu_1,\mu_2} \cdot \expt[X \mid \mu = \mu_1]$.
    \end{lemma}
    \begin{proof}
    Conceptually, generate an infinite sequence $Y_1,Y_2,\ldots$ of i.i.d.\ samples before running the algorithm. The algorithm observes only the first $X$ samples. For $j\in\set{1,2}$, define
    \myeqn{
        \varphi_j(1)
        =
        \mu_j
        \qquad\text{and}\qquad
        \varphi_j(-1)
        =
        1-\mu_j. \nn
    }
    Thus, $\varphi_j(y)$ is the probability of obtaining the sample value $y$ when $\mu=\mu_j$.

    \vgap

    Consider an arbitrary transcript $T\in\TTT$ written as
    \myeqn{
        T
        =
        \Sigma_0,(y_1,\Sigma_1),\ldots,(y_x,\Sigma_x), \nn
    }
    where $x$ is the number of samples in $T$ and $y_i\in\set{-1,1}$ is the $i$-th sample. Let $b(T)$ be the total number of random bits in $\Sigma_0,\ldots,\Sigma_x$; each such bit is fair and independent of $\mu$. Furthermore, after the random bits and sample values in $T$ have been fixed, every decision made by the algorithm --- including whether to draw another sample or to stop --- is deterministic. It follows that
    \myeqn{
        \Q_j(T)
        &=&
        2^{-b(T)}
        \prod_{i=1}^{x}\varphi_j(y_i). \nn
    }
    Hence,
    \myeqn{
        \ln\fr{\Q_1(T)}{\Q_2(T)}
        &=&
        \sum_{i=1}^{x}
        \ln\fr{\varphi_1(y_i)}{\varphi_2(y_i)}. \nn
    }

    For every fixed transcript $T\in\TTT$, let $x(T)$ be its number of samples and let
$y_i(T)$ be its $i$-th sample. Let $T^*$ denote the random transcript generated by the algorithm. By the definition of $\Q_1$,
\myeqn{
    \Q_1(T)
    &=&
    \Pr[T^*=T\mid\mu=\mu_1]. \nn
}
Furthermore, $x(T^*)=X$ and $y_i(T^*)=Y_i$. Using the definition
of KL divergence, we
obtain
\myeqn{
    D_{\mit{KL}}(\Q_1\Vert\Q_2)
    &=&
    \sum_{T\in\TTT}
    \Q_1(T)\ln\fr{\Q_1(T)}{\Q_2(T)} \nn \\
    &=&
    \sum_{T\in\TTT}
    \Pr[T^*=T\mid\mu=\mu_1]
    \sum_{i=1}^{x(T)}
    \ln\fr{\varphi_1(y_i(T))}{\varphi_2(y_i(T))} \nn \\
    &=&
    \expt\left[
        \sum_{i=1}^{x(T^*)}
        \ln\fr{\varphi_1(y_i(T^*))}
                   {\varphi_2(y_i(T^*))}
        \bigmid\mu=\mu_1
    \right] \nn \\
    &=&
    \expt\left[
        \sum_{i=1}^{X}
        \ln\fr{\varphi_1(Y_i)}{\varphi_2(Y_i)}
        \bigmid\mu=\mu_1
    \right]. \label{eqn:lb-eps-KL-Q1Q2-1}
}

    For each $i\ge1$, the event $X\ge i$ is determined by the internal randomness of the algorithm and the samples $Y_1,\ldots,Y_{i-1}$. Hence, conditioned on $\mu=\mu_1$, this event is independent of $Y_i$, which has distribution $\operatorname{Bern}(\mu_1)$. Therefore,
    \myeqn{
        \expt\left[
            \mathbbm{1}_{X\ge i}\cdot
            \ln\fr{\varphi_1(Y_i)}{\varphi_2(Y_i)}
            \bigmid \mu=\mu_1
        \right]
        &=&
        \Pr[X\ge i\mid\mu=\mu_1]
        \sum_{y\in\set{-1,1}}
        \varphi_1(y)\ln\fr{\varphi_1(y)}{\varphi_2(y)} \nn \\
        &=&
        \Pr[X\ge i\mid\mu=\mu_1]\cdot d_{\mu_1,\mu_2}. \nn
    }

    Under the premise of Lemma~\ref{lmm:lb-eps-any:distinguish}, $\expt[X]<M$ and hence $\expt[X\mid\mu=\mu_1]\le 2\expt[X]<\infty$. Furthermore, $\ln(\varphi_1(Y_i)/\varphi_2(Y_i))$ takes only two finite values. Fubini's theorem
    \footnote{Fubini's theorem states that, for any sequence of random variables $Z_1,Z_2,\ldots$ satisfying $\sum_{i\ge 1}\expt[|Z_i|]<\infty$, we have $\expt[\sum_{i\ge 1}Z_i]=\sum_{i\ge 1}\expt[Z_i]$.}
    therefore permits exchanging the expectation and summation in \eqref{eqn:lb-eps-KL-Q1Q2-1}. We can thus derive
    \myeqn{
        D_{\mit{KL}}(\Q_1\Vert\Q_2)
        &=&
        \sum_{i\ge1}
        \expt\left[
            \mathbbm{1}_{X\ge i}\cdot
            \ln\fr{\varphi_1(Y_i)}{\varphi_2(Y_i)}
            \bigmid \mu=\mu_1
        \right] \nn \\
        &=&
        d_{\mu_1,\mu_2}
        \sum_{i\ge1}\Pr[X\ge i\mid\mu=\mu_1] \nn \\
        &=&
        d_{\mu_1,\mu_2}\cdot\expt[X\mid\mu=\mu_1] \nn
    }
    where the last equality used the tail-sum identity.
\end{proof}

    The same argument with the roles of $\mu_1$ and $\mu_2$ exchanged gives
    \myeqn{
        D_{\mit{KL}}(\Q_2 \Vert \Q_1)
        &=&
        d_{\mu_1,\mu_2} \cdot \expt[X \mid \mu=\mu_2]. \nn
    }

    Suppose that the algorithm is wrong with probability at most $1/3$. Let $\EEE \subseteq \TTT$ be the event consisting of the transcripts under which the algorithm outputs $\mu_1$, and let $\overline{\EEE}=\TTT\setm\EEE$. Under the uniform prior on $\set{\mu_1,\mu_2}$, the error probability of the algorithm is
    \myeqn{
        \fr{1}{2}\left(\Q_1(\overline{\EEE})+\Q_2(\EEE)\right)
        =
        \fr{1}{2}\left(1-\left(\Q_1(\EEE)-\Q_2(\EEE)\right)\right). \nn
    }
    From \eqref{eqn:lb-eps-any:distribution:tv-def}, we know
    $\Q_1(\EEE)-\Q_2(\EEE) \le d_{\mit{TV}}(\Q_1,\Q_2)$. Hence, the error probability is at least $\fr{1}{2}(1-d_{\mit{TV}}(\Q_1,\Q_2))$. Since it is at most $1/3$, we must have $d_{\mit{TV}}(\Q_1,\Q_2) \ge 1/3$. By Pinsker's inequality\footnote{Here, $D_{\mit{KL}}$ is defined using the natural logarithm. Under this convention, Pinsker's inequality states that $d_{\mit{TV}}(\P,\Q) \le \sqrt{D_{\mit{KL}}(\P \Vert \Q)/2}$ for any two probability distributions $\P$ and $\Q$.},
    \myeqn{
        d_{\mu_1,\mu_2} \cdot \expt[X]
        =
        \fr{D_{\mit{KL}}(\Q_1 \Vert \Q_2)+D_{\mit{KL}}(\Q_2 \Vert \Q_1)}{2}
        \ge
        2d_{\mit{TV}}(\Q_1,\Q_2)^2
        \ge
        \fr{2}{9}. \nn
    }
    Finally,
    \myeqn{
        \ln \fr{1+\gamma}{1-\gamma}
        =
        2 \int_0^\gamma \fr{1}{1-z^2} \, dz
        \le
        \fr{2\gamma}{1-\gamma^2}, \nn
    }
    and hence \eqref{eqn:lb-eps-any:distribution:d-u1u2} gives $d_{\mu_1,\mu_2} \le 2\gamma^2/(1-\gamma^2)$. We conclude that
    \myeqn{
        \expt[X]
        \ge
        \fr{2}{9d_{\mu_1,\mu_2}}
        \ge
        \fr{1-\gamma^2}{9\gamma^2}
        =
        M \nn
    }
    where the last step used \eqref{eqn:lb-eps-any:M}. This contradicts the fact that $\expt[X] < M$.

\section{Conclusions} \label{sec:conclusion}

This article has provided a comprehensive study of monotone classification in $\real^d$ with relative \rev{approximation} guarantees, where the objective is to minimize the label-probing cost while finding a classifier whose error can be higher than the optimal monotone error $k^*$ by at most a $1 + \eps$ multiplicative factor. Our findings delineate the complexity landscape across the spectrum of $\eps$. For the exact case ($\eps = 0$), we established a lower bound of $\Omega(n)$ probes even in 1D space, where $n$ is the size of the input $P$, underscoring the hardness of achieving optimality. In the approximate regime ($\eps > 0$), we introduced two algorithms: the simple \ttt{RPE} algorithm, which achieves an expected error of at most $2k^*$ with $O(w \Log \fr{n}{w})$ probes where $w$ is the width of $P$, and an algorithm powered by a new ``relative-comparison coreset'' technique, which ensures $(1+\eps) k^*$ error w.h.p.\ at a cost of $O(\fr{w}{\eps^2} \Log \fr{n}{w} \cdot \log n)$. These are complemented by lower bounds of $\Omega(w \Log \fr{n}{(1+k^*) w})$ for constant $\eps \ge 1$ and $\Omega(w/\eps^2)$ for arbitrary $\eps > 0$, demonstrating that our algorithms are near-optimal asymptotically.

\vgap

For future work, it would be an intriguing challenge to shave off an $O(\log n)$ factor in the cost of our coreset-based algorithm. Equally challenging would be the task of proving a lower bound that grows strictly faster than $\Omega(w / \eps^2)$ for arbitrary $\eps > 0$.

\bibliographystyle{abbrv}
\bibliography{./ref}

\appendix

\section*{Appendix}

\section{Concentration Bounds} \label{app:abs-est}

Let $X_1, X_2, ..., X_t$ be $t$ independent Bernoulli random variables with $\Pr[X_i = 1] = \mu$ for each $i \in [t]$ (and hence $\Pr[X_i = 0] = 1 - \mu$). The following are two standard forms of Chernoff bounds \cite{mp95}:
\myitems{
    \item for any $\gamma \in (0, 1]$:
    \myeqn{
        \Pr\left[\left|\mu - \fr{1}{t} \sum_{i=1}^t X_i\right| \ge \gamma \mu \right]
        &\le&
        2\exp\left(-\fr{\gamma^2 t\mu}{3}\right);
        \label{eqn:chernoff1}
    }
    \item
    for any $\gamma \ge 0$:
    \myeqn{
        \Pr\left[\fr{1}{t} \sum_{i=1}^t X_i \ge (1+\gamma) \mu \right]
        &\le&
        \exp\left(- \fr{\gamma^2}{2+\gamma}t\mu\right).
        \label{eqn:chernoff2}
    }
}


\begin{lemma} \label{lmm&est-absolute}
    For any $\phi \in (0, 1]$ and $\delta \in (0, 1]$, it holds that
    \myeqn{
        \Pr\left[\Big|\mu - \fr{1}{t} \sum_{i=1}^t X_i\Big| \ge \phi \right]
        &\le&
        \delta
        \label{eqn:est-absolute}
    }
    as long as $t \ge \lc \max\{\fr{\mu}{\phi^2}, \fr{1}{\phi}\} \cdot 3 \ln \fr{2}{\delta} \rc$.
\end{lemma}

\begin{proof}
If $\mu \ge \phi$, we can derive
\myeqn{
    \Pr\left[\left|\mu - \fr{1}{t} \sum_{i=1}^t X_i\right| \ge \phi \right]
    &=&
    \Pr\left[\left|\mu - \fr{1}{t} \sum_{i=1}^t X_i\right| \ge \fr{\phi}{\mu} \cdot \mu \right]
    \nn \\
    \textrm{(by \eqref{eqn:chernoff1})}
    &\le&
   2\exp\left(-\fr{1}{3} \left(\fr{\phi}{\mu}\right)^2 t\mu \right)
   \nn
}
which is at most $\delta$ when $t = \lc \fr{3\mu}{\phi^2} \ln \fr{2}{\delta} \rc$.

\vgap

If $\mu < \phi$, we can derive
\myeqn{
    && \Pr\left[\left|\mu - \fr{1}{t} \sum_{i=1}^t X_i\right| \ge \phi \right] \nn \\
    &=&
    \Pr\left[\fr{1}{t} \sum_{i=1}^t X_i \ge \phi + \mu \right]
    =
    \Pr\left[\fr{1}{t} \sum_{i=1}^t X_i \ge \left(1 + \fr{\phi}{\mu} \right) \mu \right]
    \nn \\
    \textrm{(by \eqref{eqn:chernoff2})}
    &\le&
   \exp\left(-\fr{(\phi/\mu)^2}{2+\phi/\mu} \cdot t\mu \right) = \exp\left(\fr{t \phi^2}{2\mu + \phi} \right)
   \nn
}
which is at most $\delta$ when $t = \lc \fr{2\mu+\phi}{\phi^2} \ln \fr{1}{\delta} \rc \le \lc \fr{3}{\phi} \ln \fr{1}{\delta} \rc$.
\end{proof}

Consider the setting of Problem 1. Suppose that we want to estimate the number $x$ of elements in the input $P$ satisfying an arbitrary predicate $Q$. We can draw with replacement a set $S$ of $t = O(\fr{1}{\phi^2} \log \fr{1}{\delta})$ elements from $P$ uniformly at random. If $y$ is the number of elements in $S$ satisfying $Q$, Lemma~\ref{lmm&est-absolute} assures us that $(y/t) \cdot n$ approximates $x$ up to absolute error $\phi n$ with probability at least $1-\delta$. As a corollary, given any $h \in \mono$, we can utilize the aforementioned $S$ to estimate $\err_P(h)$ --- defined in \eqref{eqn:err-classifier} --- up to absolute error $\phi n$ by formulating $Q$ as follows: an element $p \in P$ satisfies $Q$ if and only if $\lab(p) \neq h(p)$.

\section{VC-Dimension, Disagreement Coefficient, \rev{Star Number, and Teaching Dimension} of Monotone Classifiers} \label{app:vc-dc-of-mono}

\rev{The input of Problem 1 is a multiset $P$ of points in $\real^d$ where each element $p \in P$ has a label from $\set{-1, 1}$. This section discusses the VC-dimension $\lambda$, disagreement coefficient $\theta$, star number $\starnum$, and teaching dimension $\chi$ of $\mono$ (the set of monotone classifiers) on $P$. These quantities are needed to understand the limitations of the existing active learning algorithms. We will show that all four quantities are $\Omega(w)$ in the worst case, where $w$ is the width of $P$.
}

\extraspacing {\bf VC-Dimension.} The VC-dimension $\lambda$ of $\mono$ on $P$ is the size of the largest subset $S \subseteq P$ that can be {\em shattered} by $\mono$; that is, for any function $f : S \rightarrow \set{-1, 1}$, there exists a classifier $h \in \mono$ such that $h(p) = f(p)$ for every $p \in S$.

\vgap

\rev
{We show that $\lambda$ is at least the width $w$ of $P$. Identify $w$ points $p_1, p_2, ..., p_w$ of $P$ such that no point dominates another (such points must exist due to Dilworth's Theorem \cite{d50}). The set of these points can be shattered by $\mono$. Specifically, for any function $f : S \rightarrow \set{-1, 1}$, we can find a classifier $h \in \mono$ such that $\forall i \in [w]$
\myitems{
    \item if $f(p_i) = 1$, then $h(q) = 1$ for any point $q \domeq p_i$;
    \item if $f(p_i) = -1$, then $h(q) = 1$ for any point $p_i \domeq q$.
}
The points $p_1, p_2, ..., p_w$ serve as evidence that $\lambda \ge w$.}

\extraspacing {\bf Disagreement Coefficient.} Given a set $\H \subseteq \mono$ of monotone classifiers, the {{\em disagreement region}} of $\H$ --- denoted as $\dis(\H)$ --- is the set of elements $p \in P$ such that  $h_1(p) \neq h_2(p)$ for some $h_1, h_2 \in \H$. Fix $h^*$ to be an optimal monotone classifier on $P$, namely, $\err_P(h^*) = k^*$. Given a real value $\rho \in (0, 1]$, define the {\em ball} --- denoted as $B(h^*, \rho)$ --- as the set of classifiers $h \in \mono$ such that $|\set{p \in P \mid h(p) \ne h^*(p)}| \le \rho n$, namely, $h$ disagrees with $h^*$ on at most $\rho n $ elements of $P$. The {\em disagreement coefficient} $\theta$ \cite{h14} of $\mono$ under the uniform distribution over $P$ is defined as
\begin{eqnarray}
    \theta &=& \max\left\{1, \sup_{\rho \in (\fr{k^*}{n}, 1]} \fr{|\dis(B(h^*, \rho))|}{\rho \cdot n}\right\}. \label{eqn:disagreement-coefficient}
\end{eqnarray}

Next, we construct a multiset $P$ whose $\theta$ value is at least its width $w$. Choose arbitrary integers $n$ and $w$ such that $w \ge 2$, $n > w^2$, and $n$ is a multiple of $w$. Identify $w$ distinct locations $x_1, x_2, ..., x_w$ in $\real^2$ where no location dominates another. Place $n/w$ points at each $x_i$ ($i \in [w]$), assign label 1 to all of them except for exactly one point, which is assigned label $-1$. This yields $n$ labeled points, which constitute $P$. It is clear that $P$ has width $w$ and  optimal monotone error $k^* = w$. Define $h^*$ to be the optimal monotone classifier that maps the entire $\real^2$ to 1.

\vgap

Set $\rho = 1/w$, which is greater than $k^*/n = w/n$ because $n > w^2$. The ball $B(h^*, \rho)$ includes every classifier $h \in \mono$ that differs from $h^*$ in at most $n/w$ elements of $P$. For each $i \in [w]$, define $h_i$ as the monotone classifier that maps the entire $P$ to 1 except the $n/w$ elements at $x_i$, which $h_i$ maps to $-1$. Thus, $h_i \in B(h^*, \rho)$ because $h_i$ disagrees with $h^*$ on $n/w = \rho n$ points. For every point $p \in P$, there exist distinct $j_1, j_2 \in [w]$ such that $h_{j_1}(p) \ne h_{j_2}(p)$. Specifically, suppose that $p$ is at location $x_{j_1}$; then $h_{j_1}(p) = -1$ but $h_{j_2}(p) = 1$ for any $j_2 \in [w] \setm \set{j_1}$. It follows that $\dis(B(h^*, \rho)) = P$; and hence, $\theta \ge \fr{|\dis(B(h^*, \rho))|}{\rho \cdot n} = w$.

\extraspacing \rev{{\bf Star Number.} The {\em star number} $\starnum$ \cite{hy15} of $\mono$ on $P$ is the largest integer $s$ such that there exist distinct points $p_1, p_2, ..., p_s \in P$ and distinct classifiers $h_0, h_1, ..., h_s \in P$ satisfying the requirement:
\begin{center}
   $\forall i \in [s]$, $h_0$ and $h_i$ assign the same label to all $p_1, p_2, ..., p_s$ \uline{except} $p_i$.
\end{center}
We show that $\starnum$ is at least the width $w$ of $P$. Identify $w$ points $p_1, p_2, ..., p_w$ of $P$ such that no point dominates another (such points must exist due to Dilworth's Theorem). Define
\myitems{
    \item $h_0$ as the monotone classifier that maps $\set{p_1, ..., p_w}$ to 1;
    \item for each $i \in [w]$, $h_i$ as the the monotone classifier that maps $\set{p_1, ..., p_w} \setm \set{p_i}$ to 1 but maps $p_i$ to $-1$.
}
The points $p_1, ..., p_w$ and classifiers $h_0, h_1, ..., h_w$ serve as evidence that $\starnum \ge w$.}

\extraspacing \rev{{\bf Teaching Dimension.} Fix an integer $m$, and consider an arbitrary $\Sigma \in P^m$ (i.e., $\Sigma$ is a sequence of $m$ elements from $P$, some of which may be identical). W.l.o.g., let $\Sigma = (p_1, p_2, ..., p_m)$. Define
\myeqn{
    \LL(\Sigma) &=&
    \set{
        (h(p_1), ..., h(p_m) \mid h \in \mono
    }. 
    \label{eqn:app:LL}
}
In other words, $\LL(\Sigma)$ includes all possible ways to label $\Sigma$ using monotone classifiers. Let $f$ be any function from $P$ to $\set{-1, 1}$. Define $\chi_0(f, \Sigma)$ as the smallest integer $s \in [m]$ such that we can find $s$ elements in $\Sigma$ to
\myitems{
    \item either reject $(f(p_1), ..., f(p_m)) \in \LL(\Sigma)$
    \item or identify a unique candidate $(l_1, ..., l_m) \in \LL(\Sigma)$ that may be $(f(p_1), ..., f(p_m))$.
}
Formally, the integer $s$ should satisfy the requirement that there is a subset $S \subseteq [m]$ with $|S| = s$ such that at most one $(l_1, ..., l_m) \in \LL(\Sigma)$ satisfies $l_i = f(p_i)$ on every $i \in S$.}

\vgap

\rev{
Let $\D$ be the uniform distribution over $P$. Define $\chi_1(f, m, \delta)$ be the smallest integer $s \in [m]$ such that
\myeqn{
    \Pr_{\Sigma \sim \D^m} [\chi_0(f, \Sigma) > s] &\le& \delta.
    \nn
}
Then, the {\em extended teaching dimension growth function} \cite{h07} of $\mono$ on $P$ is
\myeqn{
    \chi_2(m, \delta) &=&
    \max_f \chi_1(f, m, \delta) \nn
}
where the maximization is over all possible functions $f: P \rightarrow \set{-1, 1}$. According to \cite{h07}, the {\em teaching dimension} $\chi$ in the label complexity in \eqref{eqn:related-prev1} equals $\chi_2(n, \delta)$, where $n = |P|$ and $\delta = 1/\polylog n$.\footnote{This is obtained from Theorem 3 of \cite{h07}, which in our context sets $m$ to $\min\set{n, \Omega(\fr{1}{\nu + \xi})}$, where $\nu$ and $\xi$ are given in \eqref{eqn:related-set-nu} and \eqref{eqn:related-set-xi}, respectively. When $k^* = 0$, the value of $m$ equals $n$.}}

\vgap

\rev{Next, we construct a multiset $P$ of size $n$ and width $w$ for which $\chi_2(m, 1/2) \ge w/2$. As $\chi_2$ is non-descending in $\delta$, it will then follow that $\chi = \Omega(w)$.
}

\vgap

\rev{Choose arbitrary integers $n$ and $w$ such that $w \ge 2$, $n \ge 2w$, and $n$ is a multiple of $w$. Identify $w$ distinct locations $x_1, x_2, ..., x_w$ in $\real^2$ where no location dominates another. Place $n/w$ points at each $x_i$ ($i \in [w]$). These points constitute our $P$, which has size $n$ and width $w$. We fix $f$ to be the function that maps the entire $P$ to 1 and argue that $\chi_1(f, n, 1/2) \ge w/2$, which implies $\chi_2(n, 1/2) = w/2$.}

\vgap

\rev{
Draw $\Sigma \in \D^n$ (recall that $\D$ is the uniform distribution over $P$). If $\Sigma$ includes points from $y$ different locations, we must have $\chi_0(f, \Sigma) \ge y$. W.l.o.g., assume that $\Sigma$ includes points from locations $x_1, x_2, ..., x_y$. Monotone classifiers can map each location independently to label $-1$ or 1. We need at least $y$ points to identify the unique member of $\LL(\Sigma)$ that agrees with $f$ on the entire $\Sigma$. Thus, to prove $\chi(f, n, 1/2) \ge w/2$, it suffices to show
\myeqn{
    \Pr_{\Sigma \sim \D^m} [y > w/2] &>& 1/2
    \nn
}
or equivalently
\myeqn{
    \Pr_{\Sigma \sim \D^m} [w - y < w/2] &>& 1/2.
    \label{eqn:app-h1}
}
Note that $w-y$ is the number of locations having no point picked in $\Sigma$ --- call them the {\em deserted} locations. The probability that a specific location is deserted is $(1-1/w)^n \le  e^{-n/w} \le 1/e^2 < 0.14$. Hence, $\expt_{\Sigma \sim \D^m} [w-y] \le 0.14 w$. By Markov's inequality, $\Pr_{\Sigma \sim \D^m} [w - y \ge 0.3w] \le 0.14/0.3 < 1/2$. This implies the correctness of \eqref{eqn:app-h1}.}

\end{document}